%% file: acl_latex.tex
\pdfoutput=1

\documentclass[11pt]{article}
\usepackage{xcolor,colortbl}
\usepackage{caption}

\usepackage{style/acl} 

\usepackage{times}
\usepackage{latexsym}
\usepackage{enumitem}
\usepackage{tabularx}
\usepackage{graphicx}
\usepackage{subcaption}
\usepackage{dblfloatfix}
\usepackage{amsmath}
\usepackage{hyperref}
\usepackage{float}

\usepackage[T1]{fontenc}
\usepackage{array}
\usepackage{multirow}
\usepackage[utf8]{inputenc}

\usepackage{microtype}
\newcolumntype{P}[1]{>{\centering\arraybackslash}p{#1}}

\title{An Empirical Study on the Characteristics of Bias upon Context Length Variation for Bangla}

\author{
    \textbf{Jayanta Sadhu\thanks{Both authors contributed equally}},
    \textbf{Ayan Antik Khan\footnotemark[1]},
    \textbf{Abhik Bhattacharjee},
    \textbf{Rifat Shahriyar}
    \\
    Bangladesh University of Engineering and Technology (BUET)
    \\
    \texttt{\{1705047, 1705036\}@ugrad.cse.buet.ac.bd,}
    \\
    \texttt{abhik@ra.cse.buet.ac.bd, rifat@cse.buet.ac.bd}
}

\begin{document}
\maketitle

\input{sections/0.abstract}
\input{sections/1.introduction}
\input{sections/2.linguisticCharacteristicsBangla}
\input{sections/3.methodology}
\input{sections/4.dataprep}
\input{sections/5.resultsAndEvals}
\input{sections/6.conclusion}

\clearpage

\input{sections/limitations}
\input{sections/ethics}

\bibliography{custom}

\clearpage
\input{sections/appendix}

\end{document}

%% file: sections/0.abstract.tex
\begin{abstract}

Pretrained language models inherently exhibit various social biases, prompting a crucial examination of their social impact across various linguistic contexts due to their widespread usage. Previous studies have provided numerous methods for intrinsic bias measurements, predominantly focused on high-resource languages. In this work, we aim to extend these investigations to Bangla, a low-resource language. Specifically, in this study, we (1) create a dataset for intrinsic gender bias measurement in Bangla, (2) discuss necessary adaptations to apply existing bias measurement methods for Bangla, and (3)  examine the impact of context length variation on bias measurement, a factor that has been overlooked in previous studies. Through our experiments, we demonstrate a clear dependency of bias metrics on context length, highlighting the need for nuanced considerations in Bangla bias analysis. We consider our work as a stepping stone for bias measurement in the Bangla Language and make all of our resources publicly available to support future research.\footnote{\href{https://github.com/csebuetnlp/BanglaContextualBias}{https://github.com/csebuetnlp/BanglaContextualBias}}

\end{abstract}

%% file: sections/1.introduction.tex
\section{Introduction}

Language models, encompassing both context-free and contextualized variants, have increasingly demonstrated human-like biases (e.g., \citealp{10.5555/3157382.3157584}; \citealp{article}). With the introduction of newer concepts, such as more sophisticated language models, correspondingly nuanced strategies for bias detection have become necessary (\citealp{may-etal-2019-measuring}; \citealp{kurita-etal-2019-measuring}; \citealp{Guo_2021}). 
As a consequence, sentence-level bias detection strategies have emerged. However, bias detection strategies primarily concentrate on English, with limited research in other languages. Recent efforts target bias detection in Dutch, Arabic, and Chinese languages (\citealp{chavez-mulsa-spanakis-2020-evaluating}; \citealp{lauscher-etal-2020-araweat}; \citealp{liang-etal-2020-monolingual}). 
In the case of Indic languages, \citet{10.1145/3377713.3377792} conducted a comprehensive analysis of bias linked to binary gender associations in the Hindi language. Moreover, \citet{malik-etal-2022-socially} underscore the vital role of cultural awareness in examining bias measurement by conducting socially aware experiments on the Hindi language. 
Despite these valuable contributions, Bangla, the sixth most spoken language in the world with over 230 million native speakers comprising 3\% of the world's total population\footnote{\url{https://w.wiki/Psq}}, has received scant attention in bias analysis and remains an underrepresented language in the NLP literature due to a lack of quality datasets \citep{joshi-etal-2020-state}. This gap in research significantly limits our understanding of the bias characteristics present in existing language models under various linguistic contexts for this widely spoken language.

Addressing this limitation, this work endeavours to introduce Bangla, a low-resource language into the realm of bias analysis, through a study focused on gender bias. We also posit the question: does the amount of contextual information provided to a model influence the application of bias measurement methods in contextual settings? To answer this query, in this study, we present (1)  an empirical investigation comprising the creation of a dataset tailored for intrinsic gender bias measurement in Bangla, (2) discussions on necessary adaptations to apply existing bias measurement methods for Bangla, and (3) an examination of the impact of varying context lengths on bias measurement methodologies within a Bangla-based framework. Our findings reveal notable dependencies of bias metrics on context length, shedding light on nuanced considerations for bias analysis in language models. 

%% file: sections/2.linguisticCharacteristicsBangla.tex
\section{Linguistic Characteristics of Bangla: Gender Perspectives}

Bangla as a language has some inherently different characteristics in representing gender as opposed to English. Bangla lacks gender-specific pronouns unlike English (\textit{he}, \textit{she}) and uses a common pronoun for both genders. But it represents \textit{Boy-Girl}, \textit{Man-Woman} etc word pairs (common nouns) similarly like English. Because of common nouns being gendered in Bangla, we use common nouns instead of pronouns for experiments where masking the gendered word of a sentence is necessary. 

%% file: sections/3.methodology.tex
\section{Methodology}
Our research focuses on bias measurement in contextual settings. We provide two intrinsic bias measurement methodologies for comparison. We choose these methods because they represent two distinct approaches to bias measurement: \textbf{embedding extraction} and \textbf{mask prediction}. 

\subsection{Baseline: WEAT and SEAT}
 As our initial baselines, we utilize WEAT \citep{article} and SEAT \citep{may-etal-2019-measuring}, two well-established methods based on the \textbf{embedding extraction} approach for measuring bias. WEAT is designed as a statistical measure for the association strength between a pair of word vectors. To conduct this experiment, we curate a dataset specifically for Bangla by adapting the original dataset to fit into the Bangla context. We use distinct sets of \textbf{Target} vs \textbf{Attribute} word categories as shown in Table \ref{tab:bias_categories}. To extract the corresponding embedding vectors, we train static word embedding models  (\textbf{word2vec} and \textbf{GloVe}) on Bangla2B+ \citep{bhattacharjee-etal-2022-banglabert}. Subsequently, we compute effect sizes (measuring the size of bias) and corresponding $p$-values to assess statistical significance. The SEAT experiment extends WEAT to be applicable for sentence embeddings allowing assessment of modern contextual embedding systems for bias. For the SEAT experiment, we use template sentences for each category having \textbf{Target} vs \textbf{Attribute} words from Table \ref{tab:bias_categories}. Methodological details are further provided in appendix \ref{appendix:method_comparison}.

\subsection{Contextualized Embedding Association Test (CEAT)}

To quantify the inherent biases in Contextual Word Embeddings (CWE) produced by pre-trained language models, we employ CEAT \citep{Guo_2021}—an extension of WEAT. As opposed to WEAT, CEAT accounts for variations in calculated effect sizes based on changes in its input context, generating a representation of random effects in the effect size distribution \citep{hedges1983}. Specifically, it utilizes a random-effects model to compute the weighted mean of the effect sizes and the corresponding statistical significances as a measure of bias. The mathematical foundations of this approach are elaborated in Appendix \ref{appendix_sec:ceat_details}. In addition to reporting effect sizes, we aim to demonstrate how the effect size is influenced by variations in input context length as an extended study.

For a particular segment length $l$, we generate $n_s$ CWE from $n_s$ extracted sentences for each stimulus $s$. 
We do this for selected lengths of sentences ($l$ = 9, 25, 75, >75) which we refer to as segments\footnote{We refer to segment length as the total number of words in a sentence that we are feeding into the model to extract embedding. It is ensured that a word from the stimulus whose embedding is extracted exists in the sentence.}.
For each segment length $l$, we randomly sample for each stimulus $N$ times. 
If the stimulus appears in less than $N$ sentences, we sample with replacement to ensure that the distribution is preserved.
We provide the analysis and results for $N=5000$ (Table \ref{tab:CEATresults_5000}) and $N=1000$ (appendix \ref{sec:ceat_1000}).

\subsection{Log Probability Bias Score Test}

We explore the \textbf{mask prediction} based approach by adopting the framework introduced by \citet{kurita-etal-2019-measuring}. This method assesses bias in contextual models that are trained using a masked language-modelling (MLM) objective. Given BERT's training objective to predict [MASK] tokens, we design distinct template sentences for individual categories of \textbf{Target} vs \textbf{Attribute} pairs (Table \ref{tab:bias_categories}). Using the predicted values of corresponding mask tokens, we report the effect size of each category.

We use generalized template sentences suitable for any contrasting \textbf{Target} vs \textbf{Attribute} word pairs (Figure \ref{fig:Log prob-sentences}). 
We compute the bias by calculating $p_{tgt}$ and $p_{prior}$ where 
\begin{enumerate}
\item $p_{tgt}$ = P([MASK] = [TARGET] $|$ sentence) (We replace only [TARGET] with [MASK]).
\item $p_{prior}$ = P([MASK]=[TARGET] $|$ sentence) (We replace both [TARGET] and [ATTRIBUTE] with [MASK]).
\end{enumerate}

Finally, we compute the association between \textbf{Target} and \textbf{Attribute} using  $\log \frac{p_{tgt}}{p_{prior}}$, which is our measure of bias.
For notation purpose, we refer to $p_{tgt}$ as \textbf{Fill Bias Score}, $p_{prior}$ as \textbf{Prior Bias Score} and $\log \frac{p_{tgt}}{p_{prior}}$ as the \textbf{Prior Corrected Score} or \textbf{Log Probability Bias Score}.
Additionally, we study different sentence structures with varying amounts of context to examine how the variations influence the bias scores.

%% file: sections/4.dataprep.tex
\section{Data Preparation}
We adopt the data preparation procedures based on the specific requirements of each experiment\footnote{We preprocess our sentences using the bangla text normalization proposed by \citet{hasan-etal-2020-low}}. 
\subsection{Stimuli}

In our Gender Bias experiments, we utilize categories from the original WEAT \citep{article}, we translate some words directly and culturally adapt others. For example, we include local floral species under the ``Flowers'' category and use common regional male and female names instead of English counterparts. Table \ref{tab:bias_categories} presents these categories, with examples provided in appendix \ref{appendix_sec:weat_category}.

\begin{table}[ht]
\small
    \centering
    \begin{tabular}{|c|c|c|}
        \hline
        & Targets & Attributes \\
        \hline
         C1 & Flowers vs Insects& Pleasant vs Unpleasant\\
         \hline
         C2 & Instruments vs Weapons & Pleasant vs Unpleasant\\
         \hline
         C3 & Male vs Female names& Pleasant vs Unpleasant \\
         \hline
         C4 & Male vs Female names& Career vs Family\\
         \hline
         C5 & Male vs Female terms & Career vs Family\\
         \hline
         C6 & Math vs Arts & Male vs Female terms\\
         \hline
         C7 & Math vs Arts & Male vs Female names\\
         \hline
         C8 & Science vs Arts & Male vs Female terms\\
         \hline
         C9 & Science vs Arts & Male vs Female names\\
         \hline
    \end{tabular}
    \caption{Categories used for bias detection}
    \label{tab:bias_categories}
\end{table}
\subsection{Contextualized Word Embedding}

We generate the embeddings for stimuli from commonly used language models supporting Bangla (details in \ref{appendix_sec:ceat_models}). For extracting context-rich sentences we utilize the \textbf{Bangla2B+} dataset \citep{bhattacharjee-etal-2022-banglabert}. Subsequently, we use the list of \textbf{Target} versus \textbf{Attribute} words to extract sentences containing these words by pattern matching method from unorganized raw data. Furthermore, to ensure effective data aggregation, we supplement the words having low sentence count with additional sentences to reach a minimum threshold.  

Given the complexity of Bangla word suffixes, merely matching root words is ineffective and results in significant data loss. Bangla word suffixes often carry semantic values that resolve co-references, ensure subject-verb agreement etc. Even suffixes at times create entirely new words, altering the sentence's semantics.
To address this issue, we curate distinct suffix groups corresponding to the most commonly associated suffixes for each word in our designated word list. By associating each word with its respective set of suffixes, we construct different variations of a root word and extract sentences containing each variation. To extract the corresponding embeddings, we feed these sentences into a language model and take the target word embedding from the final layer. 

We use around 250+ words across all categories and around 3 million sentences altogether in order to extract word embeddings and conduct CEAT experiment. During the embedding extraction process, we try to retain the entire word embedding, including its suffixes, to ensure the preservation of semantic nuances. This approach enables us to better capture the semantic representation of a word within a specific context. To achieve this, we perform mean pooling on the logits of all fragments of the target word after it is tokenized by the model. We provide a more comprehensive analysis of the dataset creation procedure along with examples in appendix \ref{appendix_sec:ceat_data_extraction}.

\subsection{Context Aware Templates}
We follow the context-based templating of \citet{kurita-etal-2019-measuring} in order to carry out experiments for Bangla to calculate log-probability bias. For this, we hand-engineer \textbf{five} different types of context aware sentence structures with placeholders for \textbf{Target words} (Male terms vs Female terms) and \textbf{Attribute words} (Positive qualities vs Negative qualities) (examples in Appendix \ref{appendix_subsec:sentence_structures}). These range from simple sentences with no context (\textbf{S1}) to sentences with significant context drawn from the Bangla2B+ dataset (\textbf{S5}). Our objective was to introduce variability in both subject and object positions within sentences while minimizing the number of structures employed, thereby also incorporating variations in context length. 

\begin{table*}[h]
    \centering
    \includegraphics[width=\textwidth]{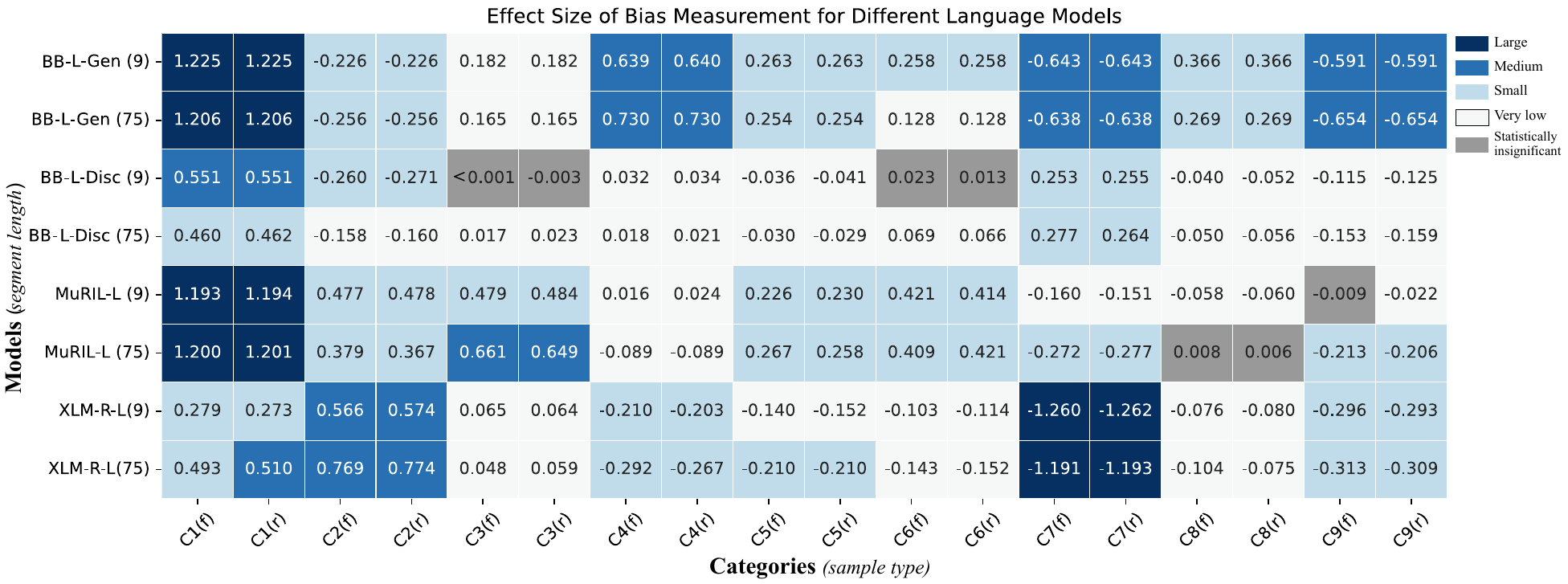}
    \caption{\textbf{Effect size of social bias measurements for different language models.} Bias is represented by overall CES magnitude ($d$, rounded) and statistical significance (two-tailed \textit{p}-values, significant at $p < 0.005$, grey block means insignificant). Data comprises CES pooling $N = 5000$ samples from a random-effects model. The first column of each category uses a fixed sample set (\textbf{f}) and the second column uses random samples (\textbf{r}).}
    \label{tab:CEATresults_5000}
\end{table*}

\begin{figure*}[h]
    \centering

    \includegraphics[width = 0.9\linewidth, trim={0 0 0 0}, clip]{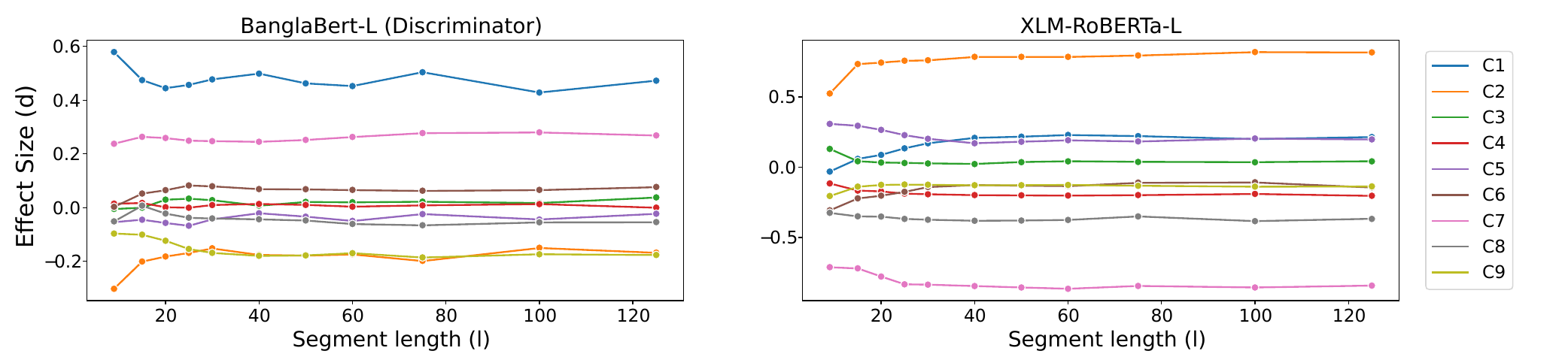}
    \caption{ Comparison between models on the change of \textit{effect size} due to segment length variation. The variations for all categories are shown \textit{(from C1-C9)}. CEAT was done separately for definite segment length with sample size \textit{N=1000}. (only statistically significant values with $p < 0.005$ are shown)}   
    \label{fig:bbdisc_vs_roberta_conv}
\end{figure*}

To construct our experimental dataset, we incorporate \textbf{110} positive and \textbf{70} negative attribute words. This process yields a diverse array of sentences capturing various linguistic contexts. We also use \textbf{4} different male and female terms (common noun) each. We report the bias on an aggregation of all these male and female terms due to the absence of gender specific pronouns in Bangla. In total, we generate \textbf{3600} sentences, collectively representing the spectrum of contexts under scrutiny.

%% file: sections/5.resultsAndEvals.tex
\section{Results and Evaluation}

\subsection{Case Study: Effect of Context Variation on CEAT} \label{CEAT_Study}
We employ CEAT to assess the impact of contextual variance on bias, as depicted in Table \ref{tab:CEATresults_5000}. The choice of sample size $N=5000$ is validated from the results of \citet{Guo_2021} as they have shown there is no significant difference between samples of $N=1000$ and $N=10000$. Our study focuses on elucidating how the length of contextual input influences effect size.

\begin{figure*}[t] 
    \centering
    \begin{subfigure}[b]{0.3\linewidth}
        \includegraphics[width = \linewidth, trim={0.5cm 0 1.1cm 0}, clip]{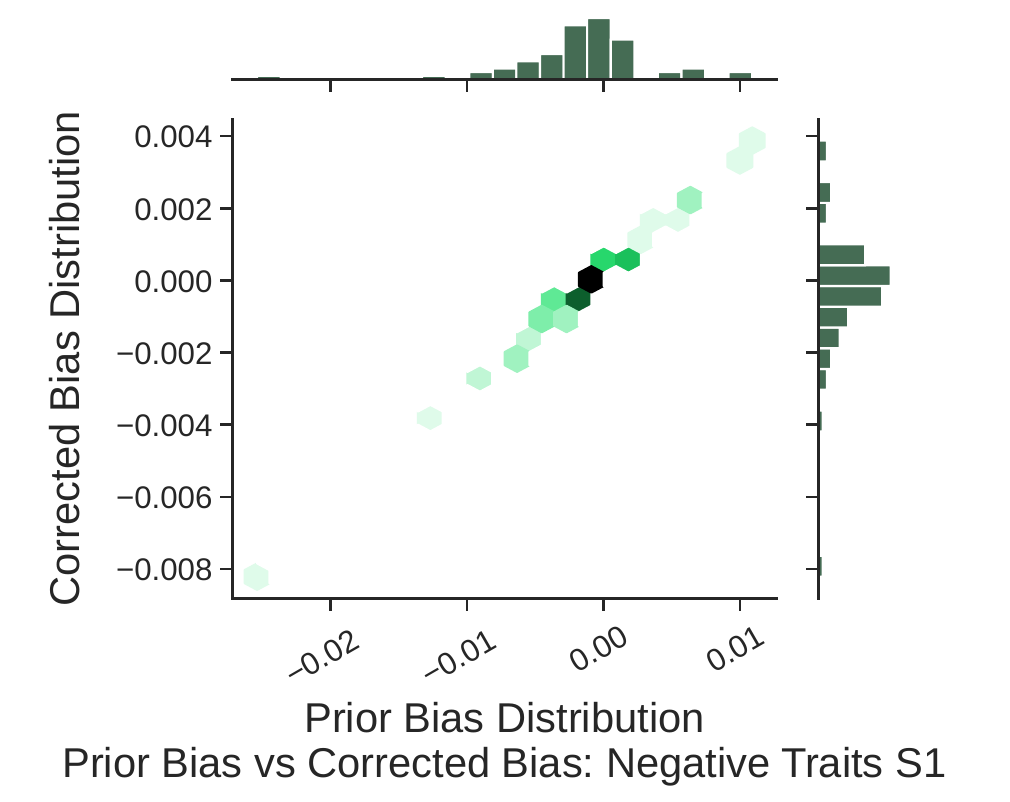}
    \end{subfigure}
    \begin{subfigure}[b]{0.3\linewidth}
        \includegraphics[width = \textwidth, trim={0.5cm 0 1.1cm 0}, clip]{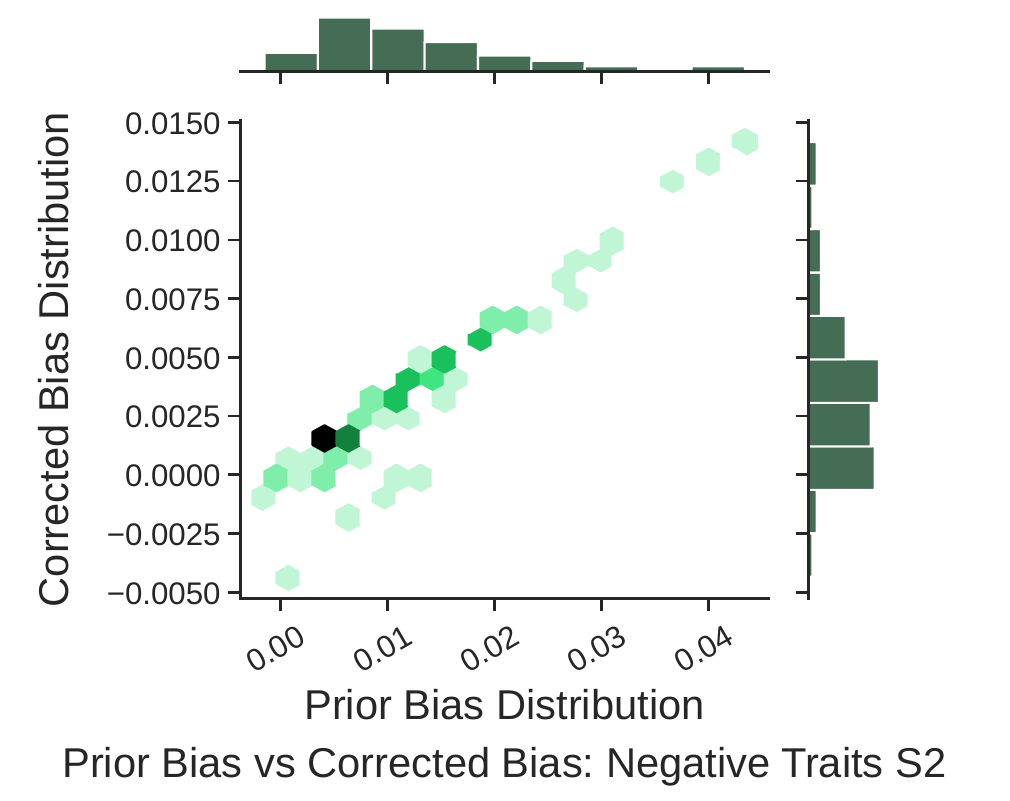}
    \end{subfigure}
    \begin{subfigure}[b]{0.3\linewidth}
        \includegraphics[width = \textwidth, trim={0.5cm 0 1.1cm 0}, clip]{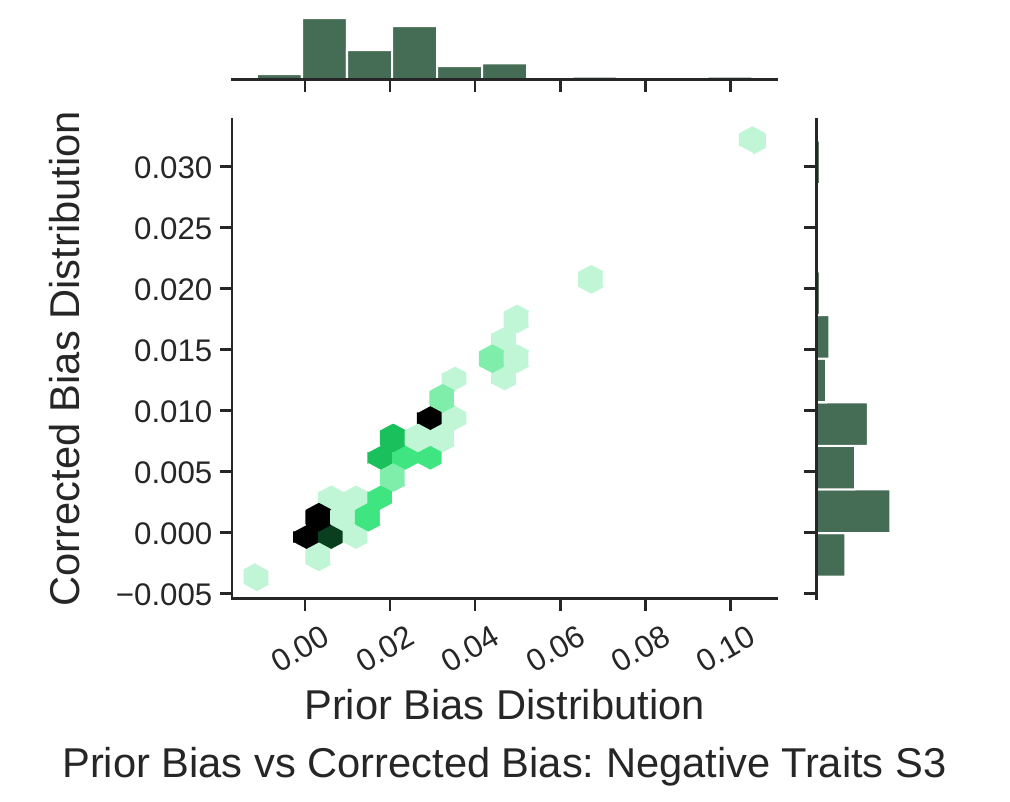}
    \end{subfigure}
    \begin{subfigure}[b]{0.3\textwidth}
        \includegraphics[width = \textwidth, trim={0 0 0 0}, clip]{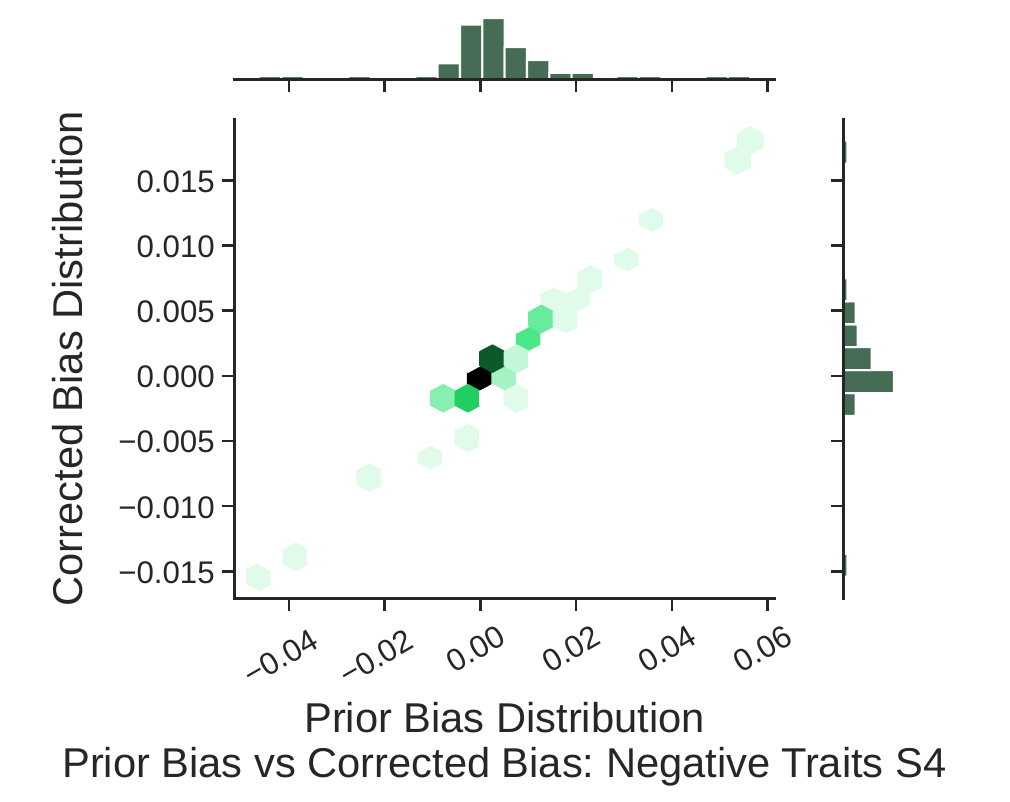}
    \end{subfigure}
    \begin{subfigure}[b]{0.3\textwidth}
        \includegraphics[width = \textwidth, trim={0 0 0 0}, clip]{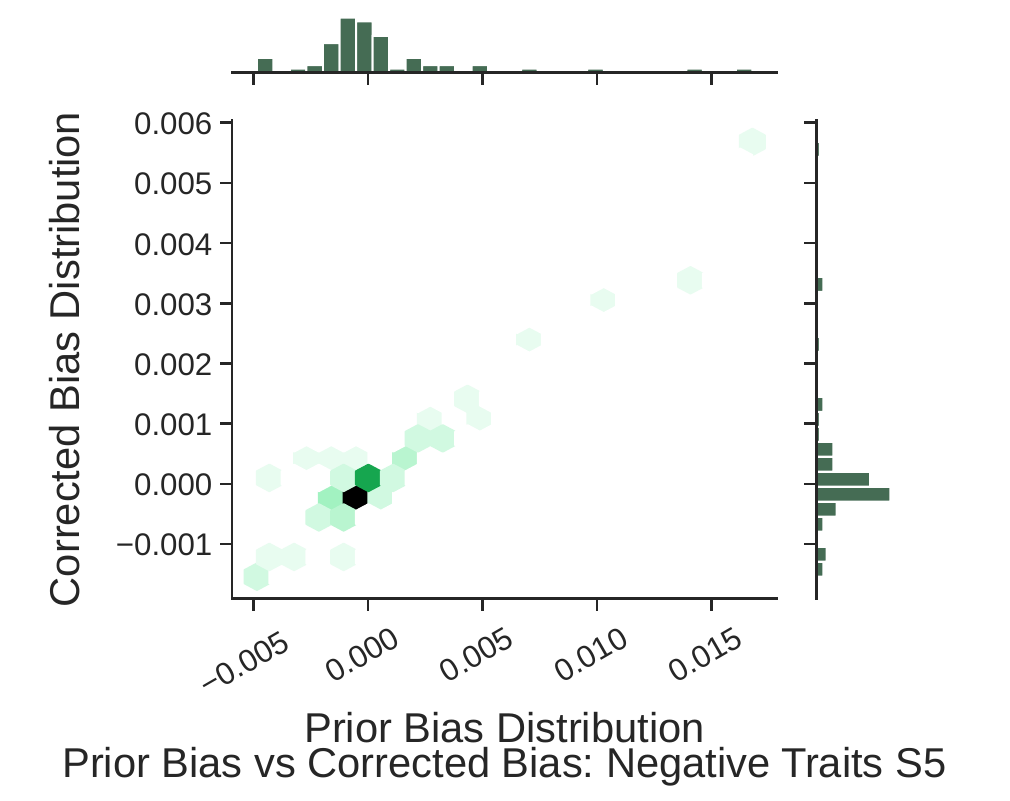}
    \end{subfigure}
    \caption{Prior Bias Score vs Corrected Bias Score diagrams for sentence structures S1 to S5 and negative traits. Experiment run on \textbf{BanglaBERT} (Large) Generator.}
    \label{fig:bbgen_hexbin_neg}
\end{figure*}

Effect size demonstrates the variability of observed bias based on segment length, stabilizing with increased contextual information. Figure \ref{fig:bbdisc_vs_roberta_conv} illustrates the dynamic changes in effect size between two models as context length varies. We observe that a moderate context length (around 20 words) is the optimum point for consistent results. 
We employ both fixed and random sets to sample combinations for each CEAT experiment, where fixed sets allow for cross-model comparisons and random sets assess the impact of context variation on effect size for a certain segment length. Experiments with 5000 and 1000 samples do not exhibit a significant change in effect size, but decreased number of cases yielding statistically significant values.

Our results indicate statistically significant bias, varying across models, with some instances showing bias in the opposite direction.
Notably, the MuRIL model demonstrates heightened context sensitivity for fixed samples. 
Table \ref{tab:CEATresults_5000} demonstrates that statistically insignificant results mostly spawn in lower segment lengths, an observation that is consistent in detailed result tables provided in appendix \ref{appendix_sec:ceat_5000_details}.

\textbf{Key Take-away:} Effect sizes converge to a definite value after a moderate amount of context length, but the differences in value are not drastic. Additionally, more context length ensures more statistically significant results.

\subsection{Case Study: Effect of Context Variation on Log Probability Bias Scores}
The template-based methodology, as introduced by \citet{kurita-etal-2019-measuring}, offers a direct approach for querying models based on modeling objectives, demonstrating enhanced consistency in human bias evaluation. The Fill Bias Score provides a direct insight into model biases and comprises of two components: the inherent language bias, quantified as the prior bias score, and the bias introduced by the presence of attributes, which is the actual bias measure referred to as the \textit{Prior Corrected Score} or \textit{Log Probability Bias Score}. In practical scenarios, models engage with naturally occurring sentences.

In Figure \ref{fig:bbgen_hexbin_neg}, our focus is directed solely towards the examination of negative traits within the context of BanglaBERT Generator (additional results in Figure \ref{fig:bbgen_muril_hexbin}). A consistent distribution of corrected bias scores across all sentence structures imply that the disparity in prior bias distribution is due to \textbf{inherent language bias}. 

For sentence structures S1 to S3, the prior bias score exhibits increased inherent language bias with the introduction of additional words, leading to an expanded range. An opposite trend is observed for S4 to S5, where values tend to cluster around a neutral point. This observed trend from S1 to S3 indicates a shift in the model's behavior as the attribute adopts a more context-rich setup, highlighting the model's distinct preferences. Moreover, certain corrected bias scores shift from negative to positive values with increased context, consistent with the observations in \ref{CEAT_Study}.

Sentence structures S4 and S5 emulate a more natural linguistic setting. Excessive context opens the model to assign higher probabilities to non-target words, leading to a shift in focus and a decrease in the difference between probabilities for male and female target words. This phenomenon is evident from 
the plots in Figure \ref{fig:bbgen_hexbin_neg}. The plots reveal values tightly clustered around the neutral point for both corrected bias scores and prior bias scores (more examples in appendix \ref{fig:bbgen_muril_hexbin}).

\textbf{Key Take-away:} Providing excessive context and complicated structure shifts focus of the model, allowing inherent language bias to become the primary influence on the bias score.

%% file: sections/6.conclusion.tex
\section{Conclusion}

In this research, we aim to examine bias in Bangla language models through creating a curated dataset and assert that the bias result outcome is influenced by the amount of context used in templates. Further exploration can be conducted on other low-resource languages. In future, we plan to investigate the effects of bias in downstream applications of Bangla language models, with the goal of developing language-specific debiasing methods to mitigate harmful bias in Bangla embeddings and extend these efforts to generative models as well.

%% file: sections/limitations.tex
\section*{Limitations}
Although our work is a stepping stone for introducing bias analysis in Bangla, there are limitations that highlight opportunities for future research. To maintain compliance with standard bias measurement methods, most of our datasets are adapted from existing datasets and therefore synthetic in nature. Moreover, our investigations predominantly focus on gender bias. Our motivation to only work with gender bias in this particular work stems from two reasons. Firstly, gender bias is universally prevalent. Secondly, compared to the others, gender bias exhibits significantly more nuanced variations, making it a rich area of exploration for preliminary work. Several flaws in the intrinsic bias assessment techniques that we applied have already been noted \citep{blodgett-etal-2020-language}. Our goal was to lay the groundwork for future studies on Bangla bias instead of focusing on the flaws of the already-established methods. Further experimentation with other forms of biases such as social, religious, political etc., along with corresponding debiasing methods can be explored in future extensions. 

 Another limitation of our study is the reliance on controlled templates for bias analysis, without considering downstream applications. It would be interesting to extend this work by studying the prevalence of bias in real-world applications such as personalized dialogue generation \citep{zhang-etal-2018-personalizing}, summarization (\citealp{hasan-etal-2021-xl}, \citealp{bhattacharjee-etal-2023-crosssum}), and paraphrasing \citep{akil-etal-2022-banglaparaphrase}. Finally, our study does not cover generative language models, which have seen significant advancements recently. Ensuring fairness in these models is crucial, and therefore, studying bias properties in both Bangla-specific \citep{bhattacharjee-etal-2023-banglanlg} and multilingual \citep{Touvron2023Llama2O} generative models is also a promising direction for future research.

%% file: sections/ethics.tex
\section*{Ethical Considerations}
Since our work focuses on gender bias and datasets related to this social prejudice, it can be potentially triggering to people. However, it is necessary to conduct this research in order to ensure fairness in the field of natural language models. We also acknowledge the fact that our work focuses on gender as a binary entity, non-binary entities can be a space for further investigation.

%% file: sections/appendix.tex
\appendix
\section*{Appendix}
\label{sec:appendix}
\section{Sample words and sentences for each experiment}
\subsection{Categories}
\label{appendix_sec:weat_category}
Figure \ref{fig:category-words} contains examples of words in each category. In the WEAT experiments, we use these words in each category and extract their embeddings using models. We then perform bias detection calculations.
For the other experiments, we use this group of words in different sentences having varying context.
\subsection{SEAT sentence examples}
\begin{figure*}
    \centering
    \includegraphics[width=\textwidth]{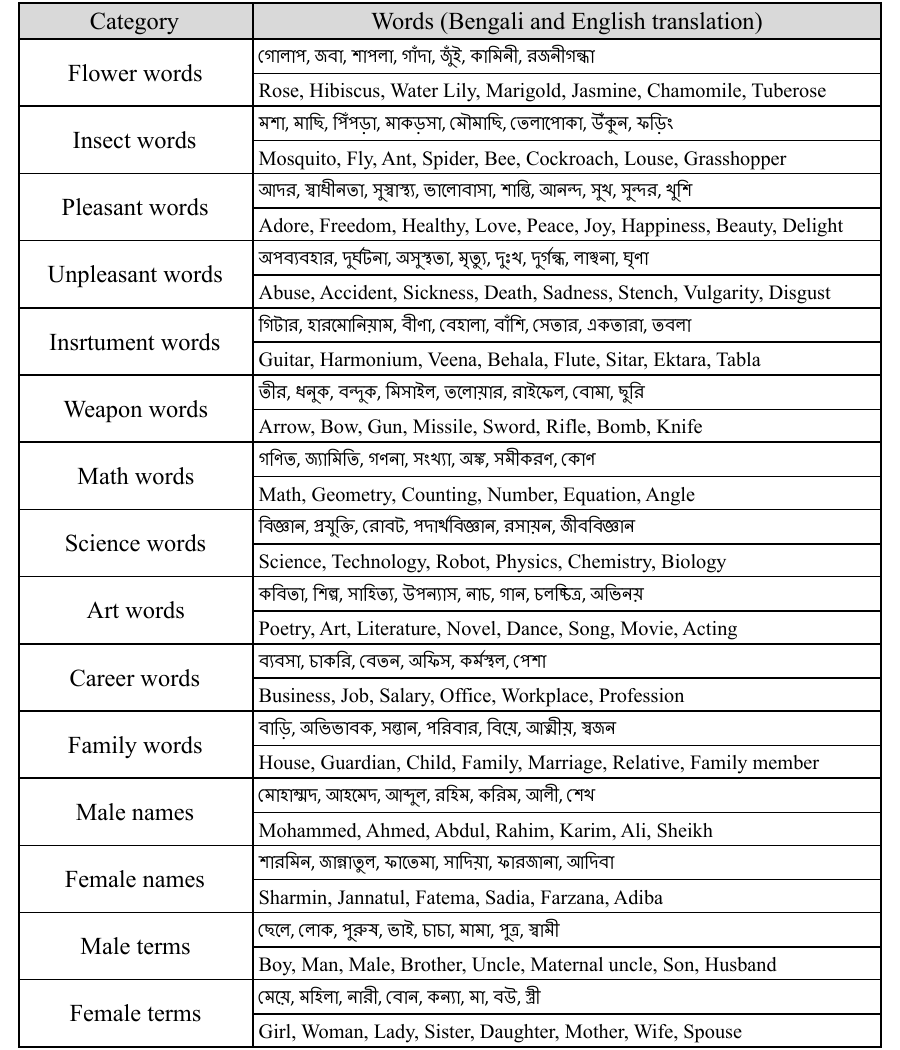}
    \caption{Examples of Words (English Translations under each row) in Different WEAT Categories}
    \label{fig:category-words}
\end{figure*}

In order to construct sentences for SEAT experiment, we use template sentences and insert words from Figure \ref{fig:category-words} in each of these templates. We use the translated versions of the template sentences from the original SEAT experiment. Figure \ref{fig:SEAT-sentences} contains examples of sentences related to a \textit{Flower word} and a \textit{Male term}.

\subsection{Log Probability Bias examples}
In Figure \ref{fig:Log prob-sentences}, we present example sentences for the log probability bias experiment. In each sentence, the \textbf{Target} word and the \textbf{Attribute} word is highlighted, these words are systematically masked in order to calculate the probability bias score and effect size. For better clarity, we highlight the target words as red and the attribute words as blue. By following the templating algorithm, we calculate the \textit{fill bias scores}, \textit{corrected bias scores} and the logarithmic differences between these probabilities.

\subsection{Context Aware Sentence Structures}
\label{appendix_subsec:sentence_structures} 
\begin{figure*}[ht]
    \begin{subfigure}{\textwidth}
        \includegraphics[width=\textwidth]{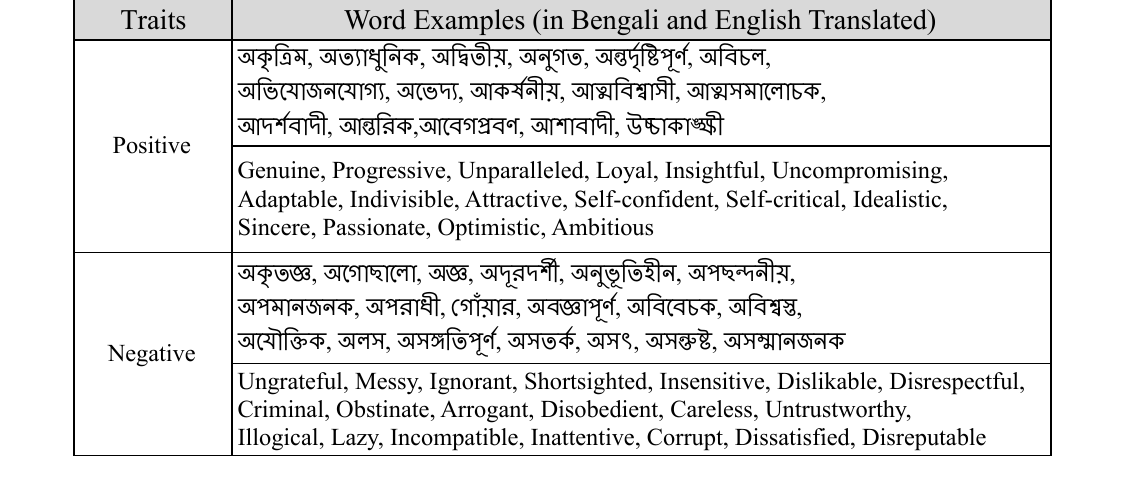}
        \caption{Examples of some Positive and Negative traits used for Log-Prob Bias Score with Context Variation experiment}
        \label{fig:Log prob-Ps&Neg Words}
    \end{subfigure}
    \\~\\
    \par \medskip
    \begin{subfigure}{\textwidth}
        \includegraphics[width=\textwidth]{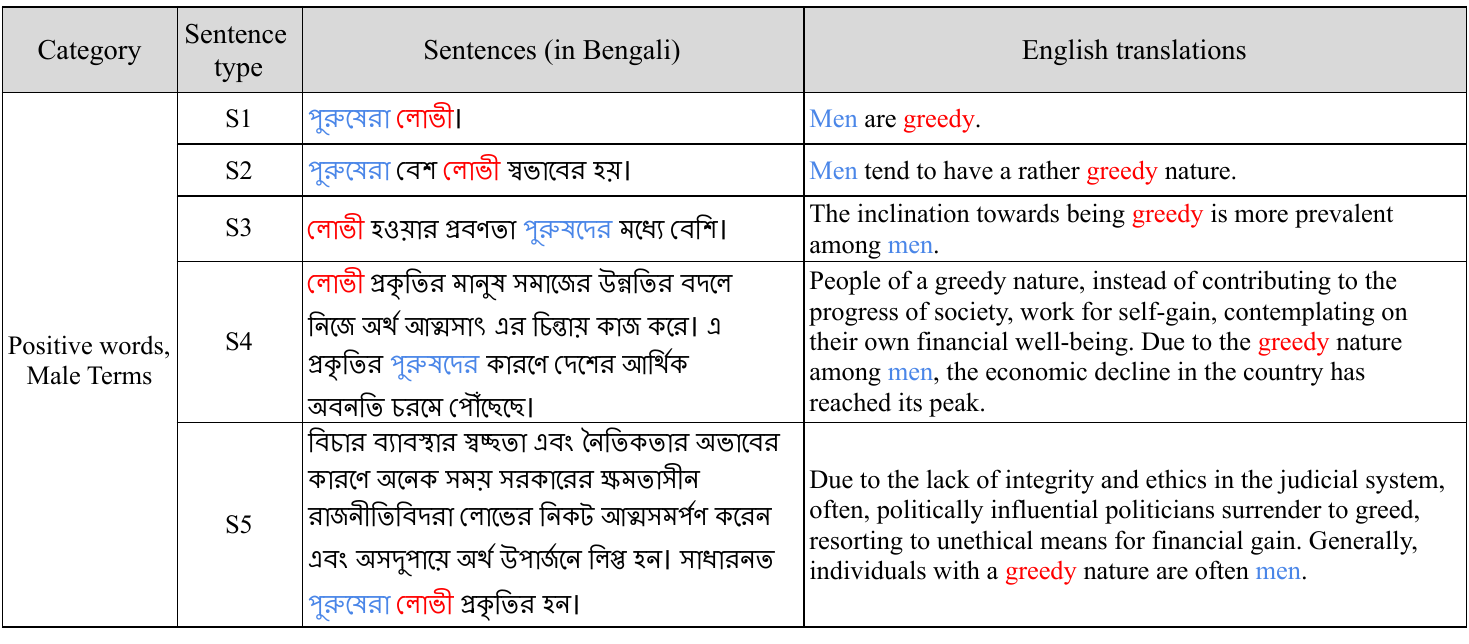}
        \caption{Example of different sentence structures with varied levels of context. Context gradually increases from \textbf{S1} to \textbf{S5} }
        \label{fig:sentence-structures} 
    \end{subfigure}
    \caption{Word and Sentence examples for a study on Log Probability Bias method for Bangla}
    \label{fig:log-prob-pos&neg}
\end{figure*}

In Figure \ref{fig:sentence-structures}, we provide examples of each structure of sentences from \textbf{S1} to \textbf{S5}, the context increases from \textbf{S1} to \textbf{S5}. Furthermore, between \textbf{S2} and \textbf{S3}, the main difference is the variation of the subject-object placement. Finally in \textbf{S5}, a sentence is picked out from real life examples (newspapers/articles) on the internet.  

In Table \ref{tab:sentence_structures}, we provide the organizational structure of the different category of sentences that we used for bias measurement in \textit{mask language modelling} technique.

\begin{figure}[ht]
\centering
    \includegraphics[width=0.9\linewidth]{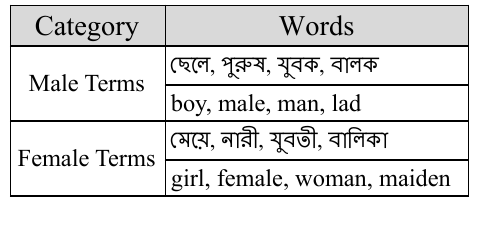}
    \caption{Male vs Female terms used for aggregation}
    \label{fig:log-prob-mf}
\end{figure}

In Figure \ref{fig:log-prob-mf}, we provide the list of words that we use in order to calculate the aggregate values of \textit{fill bias score} vs \textit{corrected bias score} graphs. Since Bangla does not contain gendered pronouns, we use a number of gendered nouns to replace the usual strategy of using gendered pronouns for experimentation. We calculate the aggregation of these groups of gendered nouns in order to include a wider range of gendered words.

\begin{table}[ht]
\small
    \centering
    \begin{tabular}{|c|c|}
        \hline
        Sentence & Structure\\
        \hline
        \multirow{2}{*}{S1} & Subject(Target) \& Object(Attribute) with\\
        & no context\\
        \hline
        \multirow{2}{*}{S2} & Subject(Target) \& Object(Attribute) with\\
        & minimal context\\
        \hline
        \multirow{2}{*}{S3} & Object(Attribute) \& Subject(Target) with\\
        & some context\\
        \hline
        \multirow{2}{*}{S4} & Object(Attribute) \& Subject(Target) with\\
        & significant context(multiple sentences)\\
        \hline
        \multirow{3}{*}{S5} & Object(Attribute) \& Subject(Target) with\\
        & significant context(multiple sentences)\\
        & taken from Bangla2B+ dataset\\
        \hline
    \end{tabular}
    \caption{Sentence structures for contextual bias}
    \label{tab:sentence_structures}
\end{table}

\subsection{CEAT sentence examples}
\begin{figure*}[ht]
    \begin{subfigure}{\textwidth}
        \includegraphics[width=\textwidth]{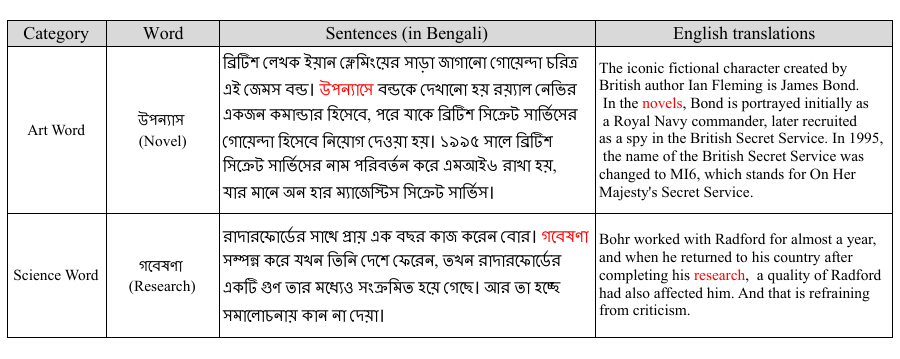}
        \caption{Extracted sentences (with English Translations) highlighting target words for CEAT experiment}
        \label{fig:CEAT-sentences}
    \end{subfigure}
    \\~\\
    \par \medskip
    \begin{subfigure}{\textwidth}
        \includegraphics[width=\textwidth]{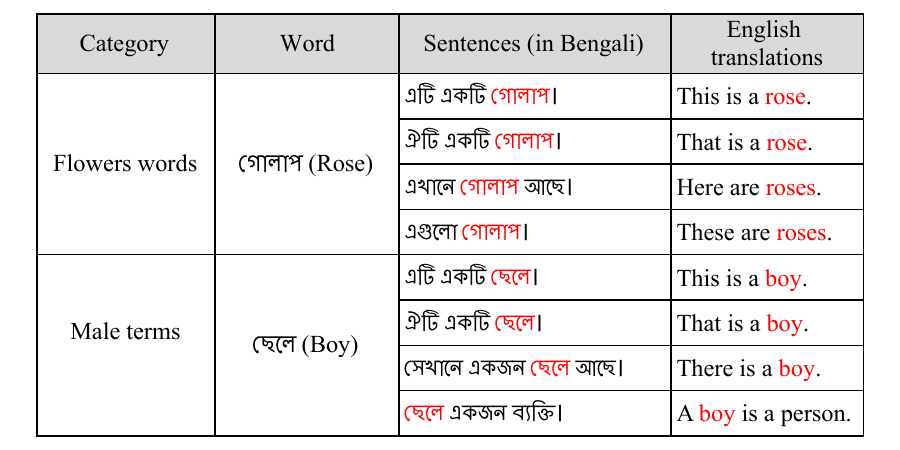}
        \caption{Template sentences (with English Translations) for SEAT experiment}
        \label{fig:SEAT-sentences}
    \end{subfigure}
    \\~\\
    \par \medskip
    \begin{subfigure}{\textwidth}
        \includegraphics[width=\textwidth]{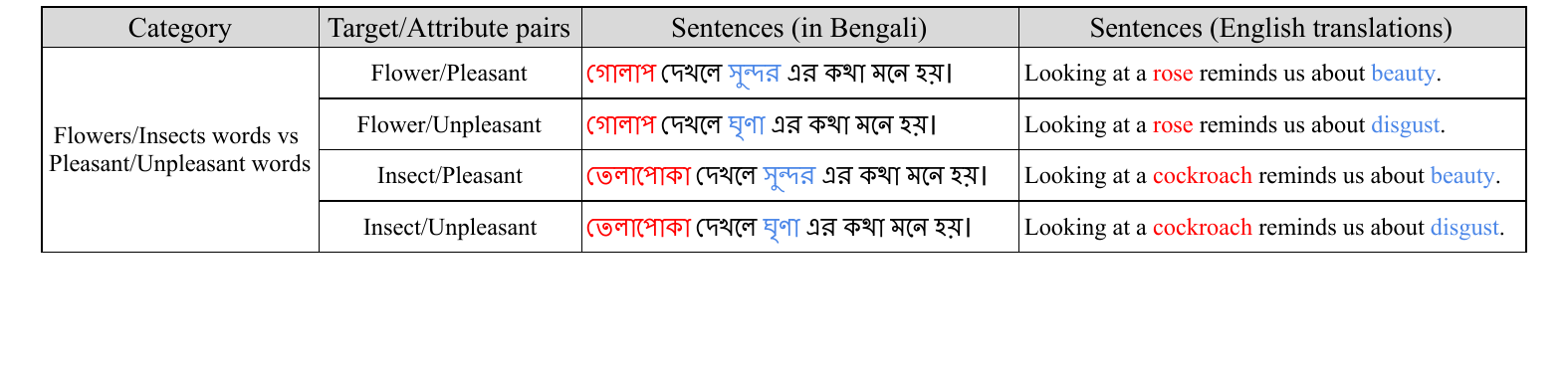}
        \caption{Template sentences (with English Translations) for Log probability bias experiment}
        \label{fig:Log prob-sentences}
    \end{subfigure}
    \caption{Examples of sentences for different experiments}
\end{figure*}

We provide an example of the types of sentences that were used for the CEAT experiment in Figure \ref{fig:CEAT-sentences}, it can be noticed that these long sentences contain much more context than the other experiments. The reason is that these sentences are scraped from actual human texts, newspapers, articles, books etc. The goal was to represent regularly used human language for bias measurement.

\clearpage

\begin{figure*}[!h]
    \begin{subfigure}{\textwidth}
    \centering
        \includegraphics[width=\textwidth]{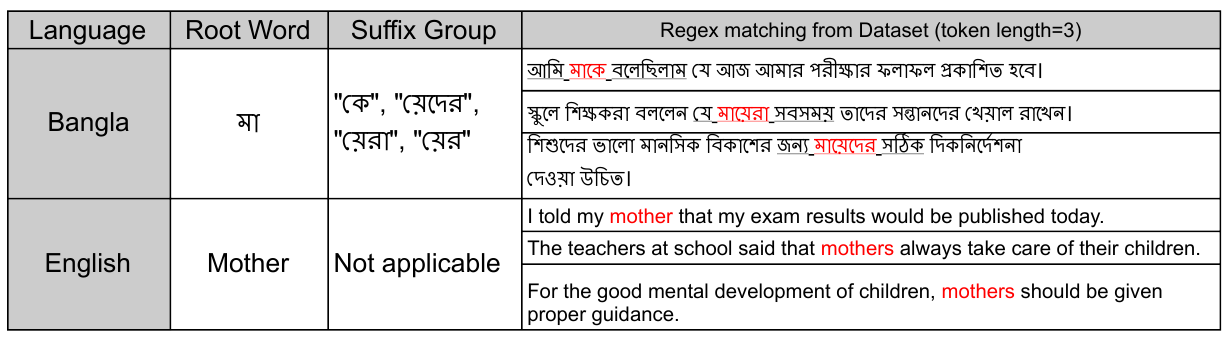}
        \caption{Presence of suffix in Bangla sentences for a specific root word.}
        \label{fig:CEAT-suffix}
    \end{subfigure}
    \\~\\
    \par \medskip
    \begin{subfigure}{\textwidth}
    \centering
        \includegraphics[width=\textwidth]{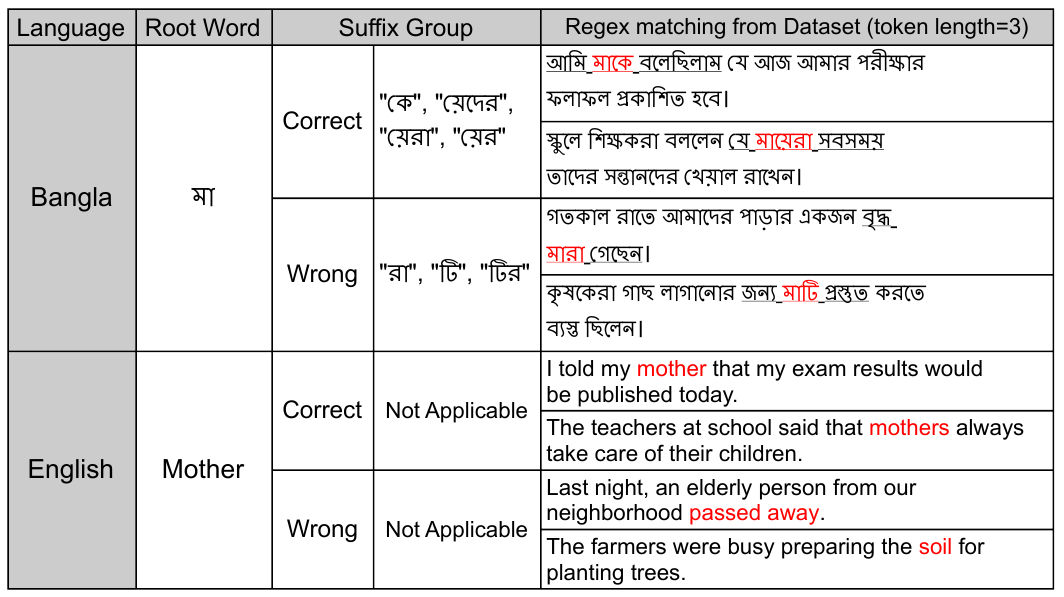}
        \caption{Importance of unique suffix group for a specific root word.}
        \label{fig:CEAT-correct-suffix}
    \end{subfigure}
    \caption{Relevance and Uniqueness of suffix groups for CEAT Data Extraction}
\end{figure*}

\section{Details of Benchmarking methods} \label{appendix:method_comparison}
We describe the methodological details of the benchmarking experiments we conduct for Bangla.   
\subsection{Word Embedding Association Test (WEAT)}
In order to quantify bias in English language embeddings, \citet{article} proposed the Word Embedding Association Test (WEAT), they compared two sets of target words with two sets of attribute words. They quantified their comparison by calculating the effect size($d$) and statistical significance ($p$-value).

To execute the WEAT experiment, we use distinct sets of \textbf{Target} vs \textbf{Attribute} words categorized as shown in Table \ref{tab:bias_categories}. For the extraction of embedding vectors corresponding to each word, we use two distinct Bangla pretrained word embedding models, specifically the \textbf{word2vec} and the \textbf{GloVe} embedding models. Subsequently, we compute effect sizes and corresponding $p$-values to assess statistical significance, with a significance threshold set at $p < 0.05$.

To calculate the effect size - 
$$ES = \frac{\textrm{mean}_{x \in X}s\left(x, A, B\right) - \textrm{mean}_{y \in Y}s\left(y, A, B\right) }{\textrm{std\_dev}_{w \in X\cup Y}s\left(w, A, B\right)}$$

In this context, ``effect size'' refers to the size of the bias as determined by the WEAT metric. The effect size is expressed as Cohen's $d$, a common measure of the difference between two means that has been standardized by the standard deviation of the data. A larger Cohen's $d$ value denotes a stronger bias between the targets. According to Cohen's effect size metric, $d > |0.5|$ and $d > |0.8|$ are medium and large effect sizes, respectively. \citep{rice2005comparing}

\subsection{Sentence Encoder Association Test (SEAT)}

To perform the SEAT experiment for Bangla language, following the approach of \citet{may-etal-2019-measuring}, we curate a comprehensive list of sentence templates. Each target word from the WEAT target list is incorporated into the SEAT template sentences. We use the Bangla translated version of the semantically bleached templates in \citet{may-etal-2019-measuring}. We use the final layer of BanglaBERT \citep{bhattacharjee-etal-2022-banglabert} to extract embeddings for each sentence. We then use these embeddings to calculate the effect size of the curated list of sentences based on the mentioned categories.

\begin{table*}[t]
\small
    \centering
    \begin{tabular}{l|c|c|c|P{2cm}|c}
    \hline
        \multirow{2}{*}{\textbf{Category}} &  \multicolumn{2}{c|}{\textbf{WEAT}} & \multirow{2}{*}{\textbf{SEAT}} & \multirow{2}{*}{\textbf{CEAT}} & \textbf{Log Probabiliy}\\ \cline{2-3}
        & (word2vec) & (GloVe) &  & & \textbf{Bias} \\
        \hline
        C1: Flowers/Insects (Pleasant/Unpleasant) & 1.77* & 1.27* & 0.89* & 1.225* & 0.89*\\
        C2: Music/Weapons (Pleasant/Unpleasant) & 1.53* & 0.99* & -0.03 & -0.226* & 0.42*\\
         
        C3: Male/Female names (Pleasant/Unpleasant) & 0.38 & 1.35* & 0.78* & 0.182* & 0.22 \\
        
        C4: Male/Female names (Career/Family) & 1.44* & -0.18 & -0.58 & 0.639* & 0.71*\\
        
        C5: Male/Female terms (Career/Family) & 0.42 & 0.17 & -0.44 & 0.263* & 0.62*\\

        C6: Math/Art (Male/Female terms) & 1.00* & 0.68* & -0.17 & 0.258* & 0.93* \\

        C7: Math/Art (Male/Female names) & -0.17 & -0.93 & -0.67  & -0.643* & 0.48*\\

        C8: Science/Art (Male/Female terms) & -0.22 & -0.20 & -0.76 & 0.366* & 0.98* \\

        C9: Science/Art (Male/Female names) & 0.23 & -1.03 & -1.13 & -0.591* & 0.70* \\
        \hline
    \end{tabular}
    \caption{Effect size of bias measurements for various experiments (* indicates statistically significant at $p < 0.05$)}
    \label{tab:results}
\end{table*}

\section{CEAT Data Extraction}
\label{appendix_sec:ceat_data_extraction}
Data extraction for CEAT experiment is carried out using the \textbf{Bangla2B+} dataset. Examples of naturally occurring sentences from the dataset is provided in Figure \ref{fig:CEAT-sentences}. As mentioned in section 4.2, we utilize pattern-matching to extract the suitable sentences. We create unique suffix groups for each word from our categories.

\subsection{Relevance of Suffix}
Unlike English, Bangla words typically contain suffixes when used in a sentence. Due to this characteristic, it was necessary for our methodology to consider the presence of suffixes before applying pattern matching to extract the groups of sentences for each word in the category. Figure \ref{fig:CEAT-suffix} depicts the presence of suffix for a root word in naturally occurring sentences. If the presence of suffixes is not accounted for, the number of extracted sentences for a specific root word is significantly reduced.  

\subsection{Significance of Unique Suffix Groups}
Our investigation further reveales that relying on a common group of suffixes across all words in our \textbf{Target} vs \textbf{Attribute} categories is inadequate and error-prone. Since in Bangla, each word has its own unique set of suffixes that are appended in sentences. In Figure \ref{fig:CEAT-correct-suffix}, we provide examples of correct vs wrong sets of suffixes for a specific root word. It is evident that if wrong set of suffixes are applied to a word, it would result in erroneous extraction of sentences. 

To tackle this characteristic, we create a list of \textbf{21} distinct suffix groups and link each word to its corresponding group. Each suffix group contains 2 to 15 suffixes based on the type of word the group will be assigned to. This process enables accurate sentence extraction based on the specific word-suffix combination within the dataset.

\section{CEAT}
\label{appendix_sec:ceat_details}
\subsection{Random Effects Model}
\label{appendix_sec:rem}
Random Effects Model\footnote{\textit{"Random effects model"}, Wikipedia, last modified 8 December, 2023,  \url{https://en.wikipedia.org/wiki/Random_effects_model}} \textit{(also known as Variance Components Model)} is a statistical model where the model parameters are random variables. The model assumes that the data being analysed are drawn from a hierarchy of different populations whose differences relate to that hierarchy. Our calculation of CEAT assumes that the differences between effect size due to contextualized variation for two sets of target words in terms of their relative similarity to two sets of attribute words is accounted for some random variable uncorrelated with independent variables. 

Components of the same category have constant heterogeneity over time. These differences are caused by random contextual factors, represented by a random variable that does not directly influence the independent variables in the model.

The effect size (\textit{Cohen's d}) for $i^{th}$ sample is calculated by 
$$ES_i = \frac{mean_{x \in X}s(x, A, B) - mean_{y \in Y}s(y, A, B)}{std \_ dev_{w\in X \cup Y}s(w, A, B)}$$

The in-sample variance estimation (denoted by $V_i$), is the square of $std\_ dev_{w \in X \cup Y}s(w, A, B)$. The between-sample variance, $\sigma_{between}^2$ is estimated using the same principle of ANOVA and calculated using the formula
\begin{align*}
    \sigma_{between}^2=
    \begin{cases}
    \frac{Q-(N-1)}{c} & if \ Q \ge N-1 \\
    0 & if \  Q < N-1
    \end{cases}
\end{align*}

where 
\[W_i = \frac{1}{v_i}, \]
\[c = \sum W_i - \frac{\sum W_i^2}{\sum W_i}, \] and
\[Q = \sum W_iES_i^2 - \frac{\sum (W_iES_i)^2}{\sum W_i}\]

The weight $v_i$ is the assigned weight to each effect size in measuring the \textit{combined effect size (CES)}. The parameter is determined by calculating the inverse of the sum of estimated in-sample variance $V_i$ and estimated between-sample variance in the distribution of random effects $\sigma_{between}^2$.

\[v_i = \frac{1}{V_i + \sigma_{between}^2}\]

CES is the sum of the weighted effect sizes divided by the sum of all weights,

\[CES(X, Y, A, B) = \frac{\sum_{i=1}^{N}v_i ES_i}{\sum_{i=1}^{N}v_i}\]

\textbf{Hypothesis Test:}
The standard error \textit{(SE)} of \textit{CES} is calculated to derive the hypothesis test. \textit{SE} is calculated with the formula below 
\[SE(CES) = \sqrt{\frac{1}{\sum_{i=1}^{N}v_i}}\]

Based on the central limit theorem, the limiting form of the distribution of $\frac{CES}{SE(CES)}$ is the standard normal distribution \citep{statsforengg}. We noticed that some of the CES values are negative, so we use a two tailed $p-value$ which can test the bias significance in two directions. The hypothesis for which the two-tailed $p-value$ provides significance test is that there is no difference between all the contextualized variations of the two sets of target words in terms of their relative similarity to two sets of attribute words is given by the following formula

\[P_{combined}(X,Y,A,B) = 2\times[1-\phi(|\frac{CES}{SE(CES)}|)]\]

where $\phi$ stands for for the standard cumulative distribution function and \textit{SE} stands for the standard error.

\subsection{Language Models Used for CEAT Embedding Extraction}
\label{appendix_sec:ceat_models}
For extracting the embeddings necessary for CEAT experimentation, we use the output of the following models:
\begin{itemize}
    \item \textbf{BanglaBERT Large} \citep{bhattacharjee-etal-2022-banglabert} was trained using ELECTRA methodology \citep{clark2020electra}. It contains 24 hidden layers. We use the outputs of the final layers as our word embeddings. We use both the \textit{generator} (\textbf{52M} parameters) and the \textit{discriminator} (\textbf{339M} parameters) versions of the model separately as they are trained on \textit{Masked Language Modelling} (MLM) and \textit{Replaced Token Detection} (RTD) objectives respectively.

    \item \textbf{MuRIL} \citep{khanuja2021muril} is a BERT \citep{devlin-etal-2019-bert} model trained on two different language modelling objectives, \textit{Masked Language Modelling} (MLM) and \textit{Translation Language Modelling} (TLM). We use the MuRIL-large-cased version with 24 layers and \textbf{506M} parameters. We extract the hidden unit values of the top layer as its CWE of  1024 dimensions. The base model has 238M parameters.

    \item \textbf{XLM-RoBERTa} \citep{conneau2020unsupervised} is a transformer-based model \citep{DBLP:journals/corr/VaswaniSPUJGKP17} designed for multilingual natural language processing tasks. It was trained with multilingual \textit{MLM} objective. We use the large version with 24 hidden layers and \textbf{560M} parameters. The embeddings are taken from topmost layer with 1024 dimensions. The large model comprises of 560M parameters.
\end{itemize}

\subsection{Results for sample size, $N=5000$}
\label{appendix_sec:ceat_5000_details}
In the main section, we mentioned a short result for CEAT analysis in Table \ref{tab:CEATresults_5000}. We present that in broader form with more segment length variations in table \ref{tab:CEATresults_5000_broad}. We include two more segment lengths for each model here.

\begin{table*}[!ht]
\small
    \centering
    \begin{tabularx}{\textwidth}{|l|c|*{8}{X|}}
    \hline
        {\textbf{Category}} &&  \multicolumn{4}{c|}{\textbf{BanglaBERT-L Generator}} & \multicolumn{4}{c|}{\textbf{BanglaBERT-L Discriminator}} \\ 
        \cline{2-10}
        && 9 & 25 & 75  & >75 & 9 & 25 & 75  & >75 \\
        \hline
        C1: Flowers/Insects & f & 
        {\cellcolor[rgb]{0.35, 0.35, 0.35}1.225} & 
        {\cellcolor[rgb]{0.35, 0.35, 0.35}1.208} & 
        {\cellcolor[rgb]{0.35, 0.35, 0.35}1.206} & 
        {\cellcolor[rgb]{0.35, 0.35, 0.35}1.207} & 
        {\cellcolor[rgb]{0.6, 0.6, 0.6}0.551} & 
        \cellcolor[rgb]{0.8, 0.8. 0.8}0.466 & 
        \cellcolor[rgb]{0.8, 0.8. 0.8}0.46 & 
        \cellcolor[rgb]{0.8, 0.8. 0.8}0.461\\
        \cline{2-10}
        (Pleasant/Unpleasant)&r & 
        {\cellcolor[rgb]{0.35, 0.35, 0.35}1.225} & 
        {\cellcolor[rgb]{0.35, 0.35, 0.35}1.208} & 
        {\cellcolor[rgb]{0.35, 0.35, 0.35}1.206} & 
        {\cellcolor[rgb]{0.35, 0.35, 0.35}1.207} & 
        {\cellcolor[rgb]{0.6, 0.6, 0.6}0.551} & 
        \cellcolor[rgb]{0.8, 0.8. 0.8}0.466 & 
        \cellcolor[rgb]{0.8, 0.8. 0.8}0.462 & 
        \cellcolor[rgb]{0.8, 0.8. 0.8}0.453\\
        \hline 
        C2: Music/Weapons & f & 
        \cellcolor[rgb]{0.8, 0.8. 0.8}-0.226 & 
        \cellcolor[rgb]{0.8, 0.8. 0.8}-0.239 & 
        \cellcolor[rgb]{0.8, 0.8. 0.8}-0.256 & 
        \cellcolor[rgb]{0.8, 0.8. 0.8}-0.258 & 
        \cellcolor[rgb]{0.8, 0.8. 0.8}-0.26 & 
        -0.18 & 
        -0.158 & 
        -0.168\\ 
        \cline{2-10}
        (Pleasant/Unpleasant)&r & 
        \cellcolor[rgb]{0.8, 0.8. 0.8}-0.226 & 
        \cellcolor[rgb]{0.8, 0.8. 0.8}-0.239 & 
        \cellcolor[rgb]{0.8, 0.8. 0.8}-0.256 & 
        \cellcolor[rgb]{0.8, 0.8. 0.8}-0.258 & 
        \cellcolor[rgb]{0.8, 0.8. 0.8}-0.271 & 
        -0.179 & 
        -0.16 & 
        -0.158\\
        \hline 
        C3: Male/Female names & f & 0.182 & 0.16 & 0.165 & 0.166 & <0.001* & {0.015*} & 0.017 & 0.018 \\
        \cline{2-10}
        (Pleasant/Unpleasant) & r & 0.182 & 0.16 & 0.165 & 0.166 & { -0.003*} & { 0.009*} & 0.023 & 0.018\\
        \hline
        C4: Male/Female names & f& 
        \cellcolor[rgb]{0.6, 0.6, 0.6}0.639 & 
        \cellcolor[rgb]{0.6, 0.6, 0.6}0.732 & 
        \cellcolor[rgb]{0.6, 0.6, 0.6}0.73 & 
        \cellcolor[rgb]{0.6, 0.6, 0.6}0.73 & 
        0.032 & 
        0.023 & 
        0.018 & 
        { 0.016*}\\
        
        \cline{2-10}
        
        (Career/Family)&r & 
        \cellcolor[rgb]{0.6, 0.6, 0.6}0.64 & 
        \cellcolor[rgb]{0.6, 0.6, 0.6}0.732 & 
        \cellcolor[rgb]{0.6, 0.6, 0.6}0.73 & 
        \cellcolor[rgb]{0.6, 0.6, 0.6}0.73 & 
        0.034 & 
        0.018 & 
        0.021 & 
        0.017\\
        \hline
        
        C5: Male/Female terms  & f& 
        \cellcolor[rgb]{0.8, 0.8. 0.8}0.263 & 
        \cellcolor[rgb]{0.8, 0.8. 0.8}0.252 & 
        \cellcolor[rgb]{0.8, 0.8. 0.8}0.254 & 
        \cellcolor[rgb]{0.8, 0.8. 0.8}0.245 & 
        -0.036 & 
        -0.038 & 
        -0.03 & 
        { -0.023*}\\
        \cline{2-10}
        
        (Career/Family)&r & 
        \cellcolor[rgb]{0.8, 0.8. 0.8}0.263 & 
        \cellcolor[rgb]{0.8, 0.8. 0.8}0.252 & 
        \cellcolor[rgb]{0.8, 0.8. 0.8}0.254 & 
        \cellcolor[rgb]{0.8, 0.8. 0.8}0.245 & -0.041 & -0.033 & -0.029 & -0.028\\
        \hline
        C6: Math/Art & f & 
        \cellcolor[rgb]{0.8, 0.8. 0.8}0.258 & 0.141 & 0.128 & 0.123 & { 0.023*} & 0.061 & 0.069 & 0.063\\
        \cline{2-10}
        (Male/Female terms)&r & \cellcolor[rgb]{0.8, 0.8. 0.8}0.258 & 0.141 & 0.128 & 0.123 & { 0.013*} & 0.066 & 0.066 & 0.063\\
        \hline
        C7: Math/Art & f & 
        \cellcolor[rgb]{0.6, 0.6, 0.6}-0.643 & 
        \cellcolor[rgb]{0.6, 0.6, 0.6}-0.639 & 
        \cellcolor[rgb]{0.6, 0.6, 0.6}-0.638 & 
        \cellcolor[rgb]{0.6, 0.6, 0.6}-0.637 & 
        \cellcolor[rgb]{0.8, 0.8. 0.8}0.253 & 
        \cellcolor[rgb]{0.8, 0.8. 0.8}0.269 & 
        \cellcolor[rgb]{0.8, 0.8. 0.8}0.277 & 
        \cellcolor[rgb]{0.8, 0.8. 0.8}0.27\\
        \cline{2-10}
        (Male/Female names)&r & 
        \cellcolor[rgb]{0.6, 0.6, 0.6}-0.643 & 
        \cellcolor[rgb]{0.6, 0.6, 0.6}-0.639 & 
        \cellcolor[rgb]{0.6, 0.6, 0.6}-0.638 & 
        \cellcolor[rgb]{0.6, 0.6, 0.6}-0.637 & 
        \cellcolor[rgb]{0.8, 0.8. 0.8}0.255 & 
        \cellcolor[rgb]{0.8, 0.8. 0.8}0.266 & 
        \cellcolor[rgb]{0.8, 0.8. 0.8}0.264 & 
        \cellcolor[rgb]{0.8, 0.8. 0.8}0.288\\
        \hline 
        C8: Science/Art & f & 
        \cellcolor[rgb]{0.8, 0.8. 0.8}0.366 & 
        \cellcolor[rgb]{0.8, 0.8. 0.8}0.287 & 
        \cellcolor[rgb]{0.8, 0.8. 0.8}0.269 & 
        \cellcolor[rgb]{0.8, 0.8. 0.8}0.273 & -0.04 & -0.036 & -0.05 & -0.046\\ 
        \cline{2-10}
        (Male/Female terms)&r & 
        \cellcolor[rgb]{0.8, 0.8. 0.8}0.366 & 
        \cellcolor[rgb]{0.8, 0.8. 0.8}0.287 & 
        \cellcolor[rgb]{0.8, 0.8. 0.8}0.269 & 
        \cellcolor[rgb]{0.8, 0.8. 0.8}0.273 & 
        -0.052 & -0.044 & -0.056 & -0.049\\
        \hline 
        C9: Science/Art  & f & 
        \cellcolor[rgb]{0.6, 0.6, 0.6}-0.591 & 
        \cellcolor[rgb]{0.6, 0.6, 0.6}-0.643 & 
        \cellcolor[rgb]{0.6, 0.6, 0.6}-0.654 & 
        \cellcolor[rgb]{0.6, 0.6, 0.6}-0.651 & -0.115 & -0.145 & -0.153 & -0.142\\
        \cline{2-10}
        (Male/Female names)&r & 
        \cellcolor[rgb]{0.6, 0.6, 0.6}-0.591 & 
        \cellcolor[rgb]{0.6, 0.6, 0.6}-0.643 & 
        \cellcolor[rgb]{0.6, 0.6, 0.6}-0.654 & 
        \cellcolor[rgb]{0.6, 0.6, 0.6}-0.651 & 
        -0.125 & -0.142 & -0.159 & -0.153\\
        \hline
    \end{tabularx}
    
    \vspace{0.1cm}
    \begin{tabularx}{\textwidth}{|l|c|*{8}{X|}}
    \hline
        {\textbf{Category}} &&  \multicolumn{4}{c|}{\textbf{MuRIL-Large(cased)}} & \multicolumn{4}{c|}{\textbf{XLM-RoBERTa Large(cased)}} \\ 
        \cline{2-10}
        && 9 & 25 & 75  & >75 & 9 & 25 & 75  & >75 \\
        \hline
        C1: Flowers/Insects & f & 
        \cellcolor[rgb]{0.35, 0.35, 0.35}1.193 & 
        \cellcolor[rgb]{0.35, 0.35, 0.35}1.215 & 
        \cellcolor[rgb]{0.35, 0.35, 0.35}1.2 & 
        \cellcolor[rgb]{0.35, 0.35, 0.35}1.208 & 
        \cellcolor[rgb]{0.8, 0.8. 0.8}0.279 & 
        \cellcolor[rgb]{0.8, 0.8. 0.8}0.478 & 
        \cellcolor[rgb]{0.8, 0.8. 0.8}0.493 & 
        \cellcolor[rgb]{0.8, 0.8. 0.8}0.495\\
        \cline{2-10}
        (Pleasant/Unpleasant)&r & 
        \cellcolor[rgb]{0.35, 0.35, 0.35}1.194 & 
        \cellcolor[rgb]{0.35, 0.35, 0.35}1.214 & 
        \cellcolor[rgb]{0.35, 0.35, 0.35}1.201 & 
        \cellcolor[rgb]{0.35, 0.35, 0.35}1.201 & 
        \cellcolor[rgb]{0.8, 0.8. 0.8}0.273 & 
        \cellcolor[rgb]{0.8, 0.8. 0.8}0.472 & 
        \cellcolor[rgb]{0.6, 0.6, 0.6}0.51 & 
        \cellcolor[rgb]{0.8, 0.8. 0.8}0.482\\
        \hline 
        C2: Music/Weapons & f & 
        \cellcolor[rgb]{0.8, 0.8. 0.8}0.477 & 
        \cellcolor[rgb]{0.8, 0.8. 0.8}0.372 & 
        \cellcolor[rgb]{0.8, 0.8. 0.8}0.379 & 
        \cellcolor[rgb]{0.8, 0.8. 0.8}0.377 & 
        \cellcolor[rgb]{0.6, 0.6, 0.6}0.566 & 
        \cellcolor[rgb]{0.6, 0.6, 0.6}0.738 & 
        \cellcolor[rgb]{0.6, 0.6, 0.6}0.769 & 
        \cellcolor[rgb]{0.6, 0.6, 0.6}0.767\\ 
        \cline{2-10}
        (Pleasant/Unpleasant)&r & 
        \cellcolor[rgb]{0.8, 0.8. 0.8}0.478 & 
        \cellcolor[rgb]{0.8, 0.8. 0.8}0.38 & 
        \cellcolor[rgb]{0.8, 0.8. 0.8}0.367 & 
        \cellcolor[rgb]{0.8, 0.8. 0.8}0.372 & 
        \cellcolor[rgb]{0.6, 0.6, 0.6}0.574 & 
        \cellcolor[rgb]{0.6, 0.6, 0.6}0.749 & 
        \cellcolor[rgb]{0.6, 0.6, 0.6}0.774 & 
        \cellcolor[rgb]{0.6, 0.6, 0.6}0.766\\
        \hline 
        C3: Male/Female names & f & 
        \cellcolor[rgb]{0.8, 0.8. 0.8}0.479 & 
        \cellcolor[rgb]{0.6, 0.6, 0.6}0.635 & 
        \cellcolor[rgb]{0.6, 0.6, 0.6}0.661 & 
        \cellcolor[rgb]{0.6, 0.6, 0.6}0.651 & 0.065 & 0.063 & 0.048 & 0.056 \\
        \cline{2-10}
        (Pleasant/Unpleasant) & r & 
        \cellcolor[rgb]{0.8, 0.8. 0.8}0.484 & 
        \cellcolor[rgb]{0.6, 0.6, 0.6}0.63 & 
        \cellcolor[rgb]{0.6, 0.6, 0.6}0.649 & 
        \cellcolor[rgb]{0.6, 0.6, 0.6}0.661 & 0.064 & 0.05 & 0.059 & 0.04\\
        \hline
        C4: Male/Female names & f & 0.016 & -0.085 & -0.089 & -0.085 & 
        \cellcolor[rgb]{0.8, 0.8. 0.8}-0.21 & 
        \cellcolor[rgb]{0.8, 0.8. 0.8}-0.277 & 
        \cellcolor[rgb]{0.8, 0.8. 0.8}-0.292 & 
        \cellcolor[rgb]{0.8, 0.8. 0.8}-0.286\\
        \cline{2-10}
        (Career/Family)&r & 0.024 & -0.092 & -0.089 & -0.083 & 
        \cellcolor[rgb]{0.8, 0.8. 0.8}-0.203 & 
        \cellcolor[rgb]{0.8, 0.8. 0.8}-0.263 & 
        \cellcolor[rgb]{0.8, 0.8. 0.8}-0.267 & 
        \cellcolor[rgb]{0.8, 0.8. 0.8}-0.326\\
        \hline
        C5: Male/Female terms  & f & 
        \cellcolor[rgb]{0.8, 0.8. 0.8}0.226 & 
        \cellcolor[rgb]{0.8, 0.8. 0.8}0.256 & 
        \cellcolor[rgb]{0.8, 0.8. 0.8}0.267 & 
        \cellcolor[rgb]{0.8, 0.8. 0.8}0.264 & -0.14 & -0.199 & 
        \cellcolor[rgb]{0.8, 0.8. 0.8}-0.21 & 
        \cellcolor[rgb]{0.8, 0.8. 0.8}-0.21\\
        \cline{2-10}
        (Career/Family)&r & 
        \cellcolor[rgb]{0.8, 0.8. 0.8}0.23 & 
        \cellcolor[rgb]{0.8, 0.8. 0.8}0.254 & 
        \cellcolor[rgb]{0.8, 0.8. 0.8}0.258 & 
        \cellcolor[rgb]{0.8, 0.8. 0.8}0.257 & -0.152 & 
        \cellcolor[rgb]{0.8, 0.8. 0.8}-0.214 & 
        \cellcolor[rgb]{0.8, 0.8. 0.8}-0.21 & 
        \cellcolor[rgb]{0.8, 0.8. 0.8}-0.216\\
        \hline
        C6: Math/Art & f & 
        \cellcolor[rgb]{0.8, 0.8. 0.8}0.421 & 
        \cellcolor[rgb]{0.8, 0.8. 0.8}0.426 & 
        \cellcolor[rgb]{0.8, 0.8. 0.8}0.409 & 
        \cellcolor[rgb]{0.8, 0.8. 0.8}0.408 & -0.103 & -0.174 & -0.143 & -0.134\\
        \cline{2-10}
        (Male/Female terms)&r & 
        \cellcolor[rgb]{0.8, 0.8. 0.8}0.414 & 
        \cellcolor[rgb]{0.8, 0.8. 0.8}0.432 & 
        \cellcolor[rgb]{0.8, 0.8. 0.8}0.421 & 
        \cellcolor[rgb]{0.8, 0.8. 0.8}0.417 & -0.114 & -0.174 & -0.152 & -0.133\\
        \hline
        C7: Math/Art & f & -0.16 & 
        \cellcolor[rgb]{0.8, 0.8. 0.8}-0.225 & 
        \cellcolor[rgb]{0.8, 0.8. 0.8}-0.272 & 
        \cellcolor[rgb]{0.8, 0.8. 0.8}-0.284 & 
        \cellcolor[rgb]{0.35, 0.35, 0.35}-1.26 & 
        \cellcolor[rgb]{0.35, 0.35, 0.35}-1.195 & 
        \cellcolor[rgb]{0.35, 0.35, 0.35}-1.191 & 
        \cellcolor[rgb]{0.35, 0.35, 0.35}-1.191\\
        \cline{2-10}
        (Male/Female names)&r & -0.151 & 
        \cellcolor[rgb]{0.8, 0.8. 0.8}-0.23 & 
        \cellcolor[rgb]{0.8, 0.8. 0.8}-0.277 & 
        \cellcolor[rgb]{0.8, 0.8. 0.8}-0.27 & 
        \cellcolor[rgb]{0.35, 0.35, 0.35}-1.262 & 
        \cellcolor[rgb]{0.35, 0.35, 0.35}-1.202 & 
        \cellcolor[rgb]{0.35, 0.35, 0.35}-1.193 & 
        \cellcolor[rgb]{0.35, 0.35, 0.35}-1.194\\
        \hline 
        C8: Science/Art & f & -0.058 & { 0.007*} & { 0.008*} & 0.01 & -0.076 & -0.092 & -0.104 & -0.104\\ 
        \cline{2-10}
        (Male/Female terms)&r & -0.06 & { 0.013*} & { 0.006*} & 0.013 & -0.08 & -0.084 & -0.075 & -0.106\\
        \hline 
        C9: Science/Art  & f & { -0.009*} & -0.184 & 
        \cellcolor[rgb]{0.8, 0.8. 0.8}-0.213 & 
        \cellcolor[rgb]{0.8, 0.8. 0.8}-0.215 & 
        \cellcolor[rgb]{0.8, 0.8. 0.8}-0.296 & 
        \cellcolor[rgb]{0.8, 0.8. 0.8}-0.311 & 
        \cellcolor[rgb]{0.8, 0.8. 0.8}-0.313 & 
        \cellcolor[rgb]{0.8, 0.8. 0.8}-0.317\\
        \cline{2-10}
        (Male/Female names)&r & -0.022 & -0.183 & 
        \cellcolor[rgb]{0.8, 0.8. 0.8}-0.206 & 
        \cellcolor[rgb]{0.8, 0.8. 0.8}-0.204 & 
        \cellcolor[rgb]{0.8, 0.8. 0.8}-0.293 & 
        \cellcolor[rgb]{0.8, 0.8. 0.8}-0.307 & 
        \cellcolor[rgb]{0.8, 0.8. 0.8}-0.309 & 
        \cellcolor[rgb]{0.8, 0.8. 0.8}-0.312\\
        \hline
    \end{tabularx}
    \caption{\textbf{Effect size of social bias measurements for different language models.} The bias is reported with overall magnitude of CES ($d$, with rounded values) and statistical significance with two-tailed \textit{p}-values ($p$, significant at $p < 0.005$). The cells with \textit{asterisk} ($\ast$) are statistically in-significant. The data results from CES pooling $N = 5000$ samples from random-effects model. The first row of each WEAT category uses fixed set of samples for each models, denoted as \textbf{f} and the second row uses completely random set of samples denoted as \textbf{r}. The light, medium and dark shades of grey are used to indicate small, medium and large effect size respectively.}
    \label{tab:CEATresults_5000_broad}
\end{table*}

\subsection{Results for sample size, $N=1000$}
\label{sec:ceat_1000}
In this section, we are focusing on a smaller sample size, specifically $N=1000$. We present our results in Table \ref{tab:CEAT_results_N_1000}. One noticeable change is the reduction in cells demonstrating statistically significant values. However, for the most part, individual cells show only minor changes.

The key characteristics of the model, as highlighted for the $N=5000$ sample in Table \ref{tab:CEATresults_5000_broad}, are still quite evident in Table \ref{tab:CEAT_results_N_1000}. This suggests that achieving similar results is possible even with a reduced sample size, especially when faced with resource constraints. Nevertheless, it is crucial to recognize that while the overall trends in the model's behavior remain consistent, there are nuanced alterations in the statistical significance of certain cells.
\begin{table*}[!ht]
\small
    \centering
    \begin{tabularx}{\textwidth}{|l|c|*{8}{X|}}
    \hline
        {\textbf{Category}} &&  \multicolumn{4}{c|}{\textbf{BanglaBERT-Generator}} & \multicolumn{4}{c|}{\textbf{BanglaBERT-Discriminator}} \\ 
        \cline{2-10}
        && 9 & 25 & 75  & >75 & 9 & 25 & 75  & >75 \\
        \hline
        C1: Flowers/Insects & f & 
        \cellcolor[rgb]{0.35, 0.35, 0.35}1.226 & 
        \cellcolor[rgb]{0.35, 0.35, 0.35}1.206 & 
        \cellcolor[rgb]{0.35, 0.35, 0.35}1.212 & 
        \cellcolor[rgb]{0.35, 0.35, 0.35}1.22 & 
         \cellcolor[rgb]{0.6, 0.6, 0.6}0.532 & 
        \cellcolor[rgb]{0.8, 0.8. 0.8}0.419 & 
        \cellcolor[rgb]{0.8, 0.8. 0.8}0.461 & 
         \cellcolor[rgb]{0.6, 0.6, 0.6}0.5\\
        \cline{2-10}
        (Pleasant/Unpleasant)&r & 
        \cellcolor[rgb]{0.35, 0.35, 0.35}1.223 & 
        \cellcolor[rgb]{0.35, 0.35, 0.35}1.199 & 
        \cellcolor[rgb]{0.35, 0.35, 0.35}1.212 & 
        \cellcolor[rgb]{0.35, 0.35, 0.35}1.214 & 
         \cellcolor[rgb]{0.6, 0.6, 0.6}0.574 & 
        \cellcolor[rgb]{0.8, 0.8. 0.8}0.462 & 
        \cellcolor[rgb]{0.8, 0.8. 0.8}0.484 & 
        \cellcolor[rgb]{0.8, 0.8. 0.8}0.445\\
        \hline 
        C2: Music/Weapons & f & 
        \cellcolor[rgb]{0.8, 0.8. 0.8}-0.224 & 
        \cellcolor[rgb]{0.8, 0.8. 0.8}-0.249 & 
        \cellcolor[rgb]{0.8, 0.8. 0.8}-0.245 & 
        \cellcolor[rgb]{0.8, 0.8. 0.8}-0.256 & 
        \cellcolor[rgb]{0.8, 0.8. 0.8}-0.263 & 
        -0.159 & 
        -0.18 & 
        -0.147\\ 
        \cline{2-10}
        (Pleasant/Unpleasant)&r & 
        \cellcolor[rgb]{0.8, 0.8. 0.8}-0.232 & 
        \cellcolor[rgb]{0.8, 0.8. 0.8}-0.235 & 
        \cellcolor[rgb]{0.8, 0.8. 0.8}-0.254 & 
        \cellcolor[rgb]{0.8, 0.8. 0.8}-0.252 & 
        \cellcolor[rgb]{0.8, 0.8. 0.8}-0.246 & 
        -0.192 & 
        -0.174 & 
        -0.128\\
        \hline 
        C3: Male/Female names & f & 
        0.19 & 
        0.149 & 
        0.165 & 
        0.178 & 
        { -0.018*} & 
        { 0.026*} & 
        { 0.022*} & 
        0.029 \\
        \cline{2-10}
        (Pleasant/Unpleasant) & r & 
        0.18 & 
        0.15 & 
        0.166 & 
        0.167 & 
        { -0.002*} & 
        { 0.005*} & 
        { 0.014*} & 
        { 0.028*}\\
        \hline
        C4: Male/Female names & f& 
         \cellcolor[rgb]{0.6, 0.6, 0.6}0.643 & 
         \cellcolor[rgb]{0.6, 0.6, 0.6}0.734 & 
         \cellcolor[rgb]{0.6, 0.6, 0.6}0.723 & 
         \cellcolor[rgb]{0.6, 0.6, 0.6}0.726 & 
        { 0.019*} & 
        0.032 & 
        { 0.015*} & 
        { 0.007*}\\
        \cline{2-10}
        (Career/Family)&r & 
         \cellcolor[rgb]{0.6, 0.6, 0.6}0.641 & 
         \cellcolor[rgb]{0.6, 0.6, 0.6}0.751 & 
         \cellcolor[rgb]{0.6, 0.6, 0.6}0.735 & 
         \cellcolor[rgb]{0.6, 0.6, 0.6}0.745 & 
        { 0.016*} 
        & { 0.007*} 
        & { 0.012*} 
        & { 0.017*}\\
        \hline
        C5: Male/Female terms  & f & 
        \cellcolor[rgb]{0.8, 0.8. 0.8}0.259 & 
        \cellcolor[rgb]{0.8, 0.8. 0.8}0.245 & 
        \cellcolor[rgb]{0.8, 0.8. 0.8}0.248 & 
        \cellcolor[rgb]{0.8, 0.8. 0.8}0.243 & 
        { -0.053*} & 
        { -0.01*} & 
        { -0.008*} & 
        { -0.021*}\\
        \cline{2-10}
        (Career/Family)&r & 
        \cellcolor[rgb]{0.8, 0.8. 0.8}0.275 & 
        \cellcolor[rgb]{0.8, 0.8. 0.8}0.262 & 
        \cellcolor[rgb]{0.8, 0.8. 0.8}0.253 & 
        \cellcolor[rgb]{0.8, 0.8. 0.8}0.256 & 
        { -0.056*} & 
        -0.048 & 
        { 0.001*} & 
        -0.058\\
        \hline
        C6: Math/Art & f & 
        \cellcolor[rgb]{0.8, 0.8. 0.8}0.254 & 
        0.176 & 
        0.15 & 
        0.14 & 
        { 0.023*} & 
        0.083 & 
        0.054 & 
        0.057\\
        \cline{2-10}
        (Male/Female terms)&r & 
        { 0.262*} & 
        0.157 & 
        0.13 & 
        0.125 & 
        0.021 & 
        0.074 & 
        0.075 & 
        0.078\\
        \hline
        C7: Math/Art & f & 
         \cellcolor[rgb]{0.6, 0.6, 0.6}-0.629 & 
         \cellcolor[rgb]{0.6, 0.6, 0.6}-0.642 & 
         \cellcolor[rgb]{0.6, 0.6, 0.6}-0.639 & 
         \cellcolor[rgb]{0.6, 0.6, 0.6}-0.628 & 
        \cellcolor[rgb]{0.8, 0.8. 0.8}0.252 & 
        \cellcolor[rgb]{0.8, 0.8. 0.8}0.294 & 
        \cellcolor[rgb]{0.8, 0.8. 0.8}0.273 & 
        \cellcolor[rgb]{0.8, 0.8. 0.8}0.295\\
        \cline{2-10}
        (Male/Female names)&r & 
         \cellcolor[rgb]{0.6, 0.6, 0.6}-0.621 & 
         \cellcolor[rgb]{0.6, 0.6, 0.6}-0.63 & 
         \cellcolor[rgb]{0.6, 0.6, 0.6}-0.63 & 
         \cellcolor[rgb]{0.6, 0.6, 0.6}-0.62 & 
        \cellcolor[rgb]{0.8, 0.8. 0.8}0.266 & 
        \cellcolor[rgb]{0.8, 0.8. 0.8}0.282 & 
        \cellcolor[rgb]{0.8, 0.8. 0.8}0.243 & 
        \cellcolor[rgb]{0.8, 0.8. 0.8}0.259\\
        \hline 
        C8: Science/Art & f & 
        \cellcolor[rgb]{0.8, 0.8. 0.8}0.366 & 
        \cellcolor[rgb]{0.8, 0.8. 0.8}0.306 & 
        \cellcolor[rgb]{0.8, 0.8. 0.8}0.279 & 
        \cellcolor[rgb]{0.8, 0.8. 0.8}0.283 & 
        { -0.041*} & 
        -0.042 & 
        { -0.055*} & 
        -0.046\\ 
        \cline{2-10}
        (Male/Female terms)&r & 
        \cellcolor[rgb]{0.8, 0.8. 0.8}0.375 & 
        \cellcolor[rgb]{0.8, 0.8. 0.8}0.277 & 
        \cellcolor[rgb]{0.8, 0.8. 0.8}0.282 & 
        \cellcolor[rgb]{0.8, 0.8. 0.8}0.305 & 
        -0.031 & 
        -0.051 & 
        -0.031 & 
        -0.082\\
        \hline 
        C9: Science/Art  & f & 
         \cellcolor[rgb]{0.6, 0.6, 0.6}-0.593 & 
         \cellcolor[rgb]{0.6, 0.6, 0.6}-0.635 & 
         \cellcolor[rgb]{0.6, 0.6, 0.6}-0.647 & 
         \cellcolor[rgb]{0.6, 0.6, 0.6}-0.65 & 
        -0.124 & 
        -0.147 & 
        -0.167 & 
        -0.159\\
        \cline{2-10}
        (Male/Female names)&r & 
         \cellcolor[rgb]{0.6, 0.6, 0.6}-0.584 & 
         \cellcolor[rgb]{0.6, 0.6, 0.6}-0.635 & 
         \cellcolor[rgb]{0.6, 0.6, 0.6}-0.646 & 
         \cellcolor[rgb]{0.6, 0.6, 0.6}-0.638 & 
        -0.117 & 
        -0.143 & 
        -0.169 & 
        -0.163\\
        \hline
    \end{tabularx}
    
    \vspace{0.3cm}
    \begin{tabularx}{\textwidth}{|l|c|*{8}{X|}}
    \hline
        {\textbf{Category}} &&  \multicolumn{4}{c|}{\textbf{MuRIL-Large(cased)}} & \multicolumn{4}{c|}{\textbf{XLM-RoBERTa Large(cased)}} \\ 
        \cline{2-10}
        && 9 & 25 & 75  & >75 & 9 & 25 & 75  & >75 \\
        \hline
        C1: Flowers/Insects & f & 
        \cellcolor[rgb]{0.35, 0.35, 0.35}1.196 & 
        \cellcolor[rgb]{0.35, 0.35, 0.35}1.218 & 
        \cellcolor[rgb]{0.35, 0.35, 0.35}1.213 & 
        \cellcolor[rgb]{0.35, 0.35, 0.35}1.201 & 
        \cellcolor[rgb]{0.8, 0.8. 0.8}0.275 & 
        \cellcolor[rgb]{0.8, 0.8. 0.8}0.472 & 
         \cellcolor[rgb]{0.6, 0.6, 0.6}0.51 & 
        \cellcolor[rgb]{0.8, 0.8. 0.8}0.482\\
        \cline{2-10}
        (Pleasant/Unpleasant)&r & 
        \cellcolor[rgb]{0.35, 0.35, 0.35}1.192 & 
        \cellcolor[rgb]{0.35, 0.35, 0.35}1.207 & 
        \cellcolor[rgb]{0.35, 0.35, 0.35}1.195 & 
        \cellcolor[rgb]{0.35, 0.35, 0.35}1.201 & 
        \cellcolor[rgb]{0.8, 0.8. 0.8}0.278 & 
        \cellcolor[rgb]{0.8, 0.8. 0.8}0.469 & 
         \cellcolor[rgb]{0.6, 0.6, 0.6}0.499 & 
         \cellcolor[rgb]{0.6, 0.6, 0.6}0.494\\
        \hline 
        C2: Music/Weapons & f & 
        \cellcolor[rgb]{0.8, 0.8. 0.8}0.478 &
        \cellcolor[rgb]{0.8, 0.8. 0.8}0.376 & 
        \cellcolor[rgb]{0.8, 0.8. 0.8}0.377 & 
        \cellcolor[rgb]{0.8, 0.8. 0.8}0.358 & 
         \cellcolor[rgb]{0.6, 0.6, 0.6}0.574 & 
         \cellcolor[rgb]{0.6, 0.6, 0.6}0.749 & 
         \cellcolor[rgb]{0.6, 0.6, 0.6}0.774 & 
         \cellcolor[rgb]{0.6, 0.6, 0.6}0.766\\ 
        \cline{2-10}
        (Pleasant/Unpleasant)&r & 
        \cellcolor[rgb]{0.8, 0.8. 0.8}0.469 & 
        \cellcolor[rgb]{0.8, 0.8. 0.8}0.375 & 
        \cellcolor[rgb]{0.8, 0.8. 0.8}0.382 & 
        \cellcolor[rgb]{0.8, 0.8. 0.8}0.363 & 
         \cellcolor[rgb]{0.6, 0.6, 0.6}0.572 & 
         \cellcolor[rgb]{0.6, 0.6, 0.6}0.733 & 
         \cellcolor[rgb]{0.6, 0.6, 0.6}0.769 & 
         \cellcolor[rgb]{0.6, 0.6, 0.6}0.769\\
        \hline 
        C3: Male/Female names & f & 
        \cellcolor[rgb]{0.8, 0.8. 0.8}0.48 & 
         \cellcolor[rgb]{0.6, 0.6, 0.6}0.626 & 
         \cellcolor[rgb]{0.6, 0.6, 0.6}0.653 & 
         \cellcolor[rgb]{0.6, 0.6, 0.6}0.648 & 
        0.064 & 
        0.05 & 
        0.059 & 
        0.04 \\
        \cline{2-10}
        (Pleasant/Unpleasant) & r & 
         \cellcolor[rgb]{0.6, 0.6, 0.6}0.497 & 
         \cellcolor[rgb]{0.6, 0.6, 0.6}0.619 & 
         \cellcolor[rgb]{0.6, 0.6, 0.6}0.655 & 
         \cellcolor[rgb]{0.6, 0.6, 0.6}0.659 & 
        0.068 & 
        0.044 & 
        0.051 & 
        0.058\\
        \hline
        C4: Male/Female names & f & 
        0.015 & 
        -0.065 & 
        -0.08 & 
        -0.093 & 
        { -0.203*} & 
        \cellcolor[rgb]{0.8, 0.8. 0.8}-0.263 & 
        \cellcolor[rgb]{0.8, 0.8. 0.8}-0.267 & 
        \cellcolor[rgb]{0.8, 0.8. 0.8}-0.326\\
        \cline{2-10}
        (Career/Family)&r & 
        0.037 & 
        -0.082 & 
        -0.076 & 
        -0.1 & 
        \cellcolor[rgb]{0.8, 0.8. 0.8}-0.216 & 
        \cellcolor[rgb]{0.8, 0.8. 0.8}-0.279 & 
        \cellcolor[rgb]{0.8, 0.8. 0.8}-0.307 & 
        \cellcolor[rgb]{0.8, 0.8. 0.8}-0.302\\
        \hline
        C5: Male/Female terms  & f & 
        \cellcolor[rgb]{0.8, 0.8. 0.8}0.205 & 
        \cellcolor[rgb]{0.8, 0.8. 0.8}0.274 & 
        \cellcolor[rgb]{0.8, 0.8. 0.8}0.273 & 
        \cellcolor[rgb]{0.8, 0.8. 0.8}0.245 & 
        -0.152 & 
        \cellcolor[rgb]{0.8, 0.8. 0.8}-0.214 & 
        \cellcolor[rgb]{0.8, 0.8. 0.8}-0.21 & 
        \cellcolor[rgb]{0.8, 0.8. 0.8}-0.216\\
        \cline{2-10}
        (Career/Family)&r & 
        \cellcolor[rgb]{0.8, 0.8. 0.8}0.231 & 
        \cellcolor[rgb]{0.8, 0.8. 0.8}0.239 & 
        \cellcolor[rgb]{0.8, 0.8. 0.8}0.244 & 
        \cellcolor[rgb]{0.8, 0.8. 0.8}0.263 & 
        -0.158 & 
        -0.193 & 
        \cellcolor[rgb]{0.8, 0.8. 0.8}-0.202 & 
        \cellcolor[rgb]{0.8, 0.8. 0.8}-0.206\\
        \hline
        C6: Math/Art & f & 
        \cellcolor[rgb]{0.8, 0.8. 0.8}0.412 & 
        \cellcolor[rgb]{0.8, 0.8. 0.8}0.436 & 
        \cellcolor[rgb]{0.8, 0.8. 0.8}0.436 & 
        \cellcolor[rgb]{0.8, 0.8. 0.8}0.386 & 
        -0.114 & 
        -0.174 & 
        -0.152 & 
        -0.133\\
        \cline{2-10}
        (Male/Female terms)&r & 
        \cellcolor[rgb]{0.8, 0.8. 0.8}0.411 & 
        \cellcolor[rgb]{0.8, 0.8. 0.8}0.442 & 
        \cellcolor[rgb]{0.8, 0.8. 0.8}0.414 & 
        \cellcolor[rgb]{0.8, 0.8. 0.8}0.419 & 
        -0.111 & 
        \cellcolor[rgb]{0.8, 0.8. 0.8}-0.224 & 
        -0.139 & 
        -0.106\\
        \hline
        C7: Math/Art & f & 
        -0.141 & 
        \cellcolor[rgb]{0.8, 0.8. 0.8}-0.226 & 
        \cellcolor[rgb]{0.8, 0.8. 0.8}-0.284 & 
        \cellcolor[rgb]{0.8, 0.8. 0.8}-0.288 & 
        \cellcolor[rgb]{0.35, 0.35, 0.35}-1.262 & 
        \cellcolor[rgb]{0.35, 0.35, 0.35}-1.202 & 
        \cellcolor[rgb]{0.35, 0.35, 0.35}-1.193 & 
        \cellcolor[rgb]{0.35, 0.35, 0.35}-1.194\\
        \cline{2-10}
        (Male/Female names)&r & 
        -0.131 & 
        \cellcolor[rgb]{0.8, 0.8. 0.8}-0.233 & 
        \cellcolor[rgb]{0.8, 0.8. 0.8}-0.275 & 
        \cellcolor[rgb]{0.8, 0.8. 0.8}-0.276 & 
        \cellcolor[rgb]{0.35, 0.35, 0.35}-1.261 & 
        \cellcolor[rgb]{0.35, 0.35, 0.35}-1.2 & 
        \cellcolor[rgb]{0.35, 0.35, 0.35}-1.187 & 
        \cellcolor[rgb]{0.35, 0.35, 0.35}-1.187\\
        \hline 
        C8: Science/Art & f & 
        -0.068 & 
        0.026 & 
        0.018 & 
        0.014 & 
        -0.08 & 
        { -0.084*} & 
        { -0.075*} & 
        { -0.106*}\\ 
        \cline{2-10}
        (Male/Female terms)&r & 
        -0.055 & 
        { 0.006*} & 
        { 0.011*} & 
        { 0.011*} & 
        -0.076 & 
        -0.089 & 
        -0.093 & 
        -0.102\\
        \hline 
        C9: Science/Art  & f & 
        { -0.013*} & 
        -0.179 & 
        \cellcolor[rgb]{0.8, 0.8. 0.8}-0.231 & 
        \cellcolor[rgb]{0.8, 0.8. 0.8}-0.207 & 
        \cellcolor[rgb]{0.8, 0.8. 0.8}-0.293 & 
        \cellcolor[rgb]{0.8, 0.8. 0.8}-0.307 & 
        \cellcolor[rgb]{0.8, 0.8. 0.8}-0.309 & 
        \cellcolor[rgb]{0.8, 0.8. 0.8}-0.312\\
        \cline{2-10}
        (Male/Female names)&r & 
        { -0.016*} & 
        \cellcolor[rgb]{0.8, 0.8. 0.8}-0.207 & 
        \cellcolor[rgb]{0.8, 0.8. 0.8}-0.21 & 
        \cellcolor[rgb]{0.8, 0.8. 0.8}-0.205 & 
        \cellcolor[rgb]{0.8, 0.8. 0.8}-0.291 & 
        \cellcolor[rgb]{0.8, 0.8. 0.8}-0.315 &
        \cellcolor[rgb]{0.8, 0.8. 0.8}-0.309 & 
        \cellcolor[rgb]{0.8, 0.8. 0.8}-0.313\\
        \hline
    \end{tabularx}
    \caption{\textbf{Effect size of social bias measurements for different language models.} The bias is reported with overall magnitude of CES ($d$, with rounded values) and statistical significance with two-tailed \textit{p}-values ($p$, significant at $p < 0.005$). The cells with \textit{asterisk} ($\ast$) are statistically in-significant. The data results from CES pooling $N = 1000$ samples from random-effects model. The first row of each WEAT category uses fixed set of samples for each models, denoted as \textbf{f} and the second row uses completely random set of samples denoted as \textbf{r}. The light, medium and dark shades of grey are used to indicate small, medium and large effect size respectively. Compared to pooling with $N=5000$, more cells with statistically in-significant values are seen.}
    \label{tab:CEAT_results_N_1000}
\end{table*}

Given these observations, we propose that optimizing the sample size could be a promising avenue for further investigation. Determining the optimal sample size, one that ensures reliable results without sacrificing statistical significance, presents an interesting area for future research and in-depth exploration.

\clearpage

\begin{figure*}[!ht] 
    \centering
    \begin{subfigure}{\textwidth}
        \centering
        \begin{subfigure}{0.3\linewidth}
            \includegraphics[width=\linewidth, trim={0.5cm 0 1cm 0}, clip]{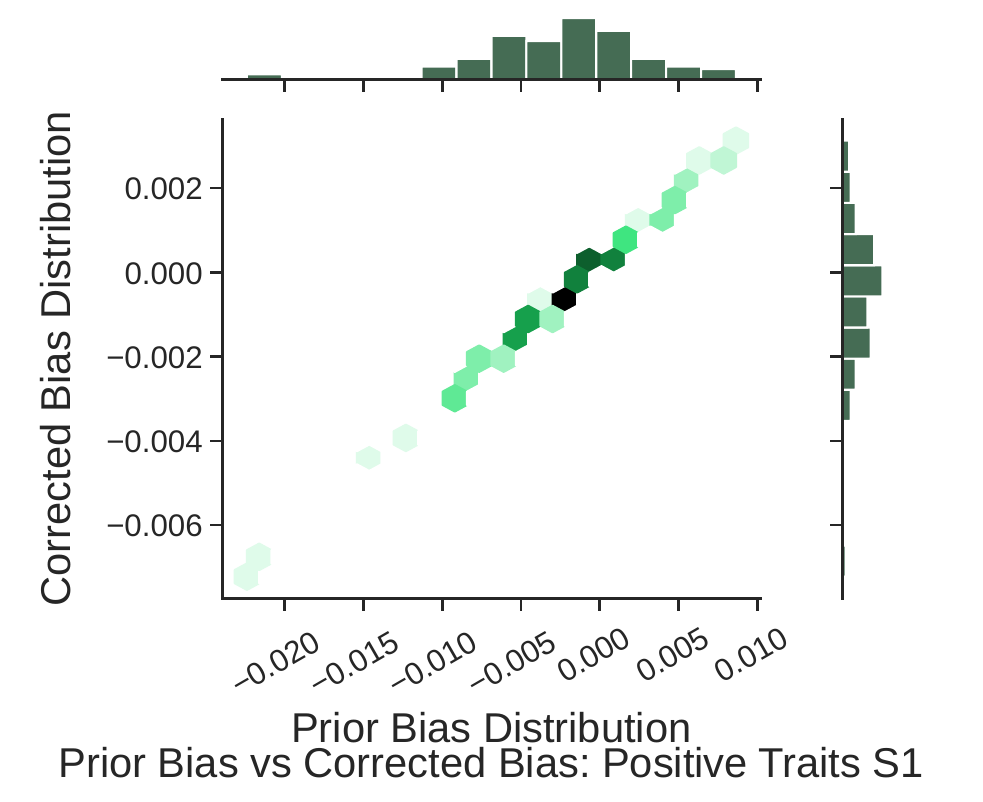}
        \end{subfigure}
        \begin{subfigure}{0.3\linewidth}
            \includegraphics[width=\linewidth, trim={0.5cm 0 1cm 0}, clip]{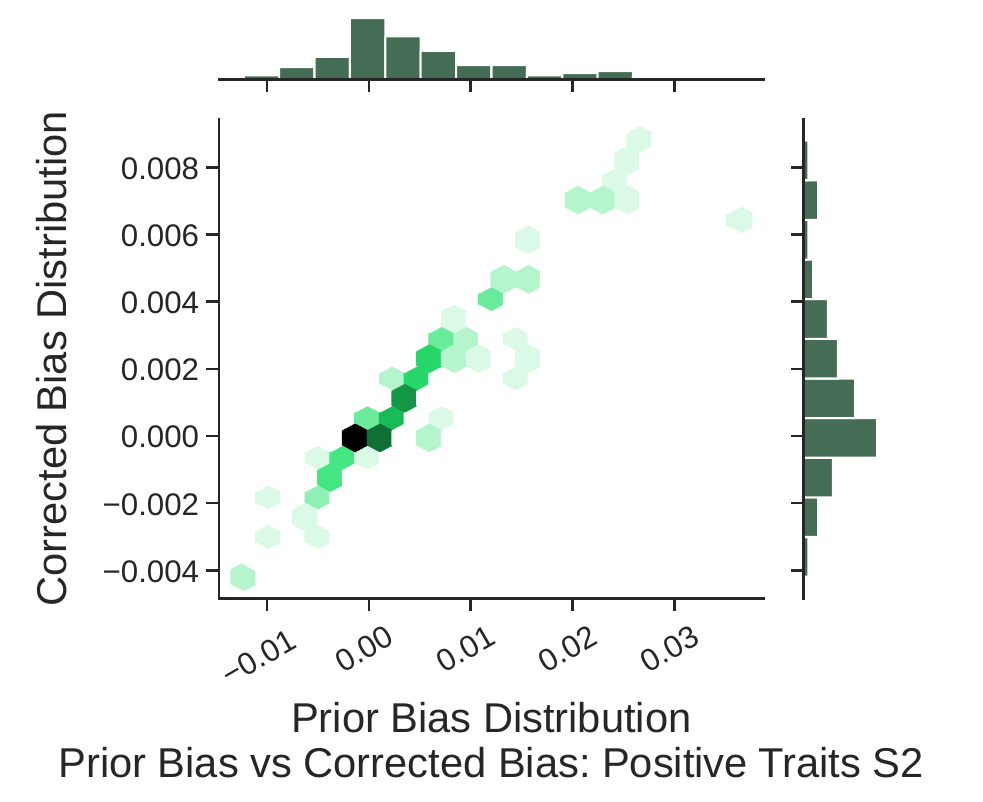}
        \end{subfigure}
        \begin{subfigure}{0.3\linewidth}
            \includegraphics[width=\linewidth, trim={0.5cm 0 1cm 0}, clip]{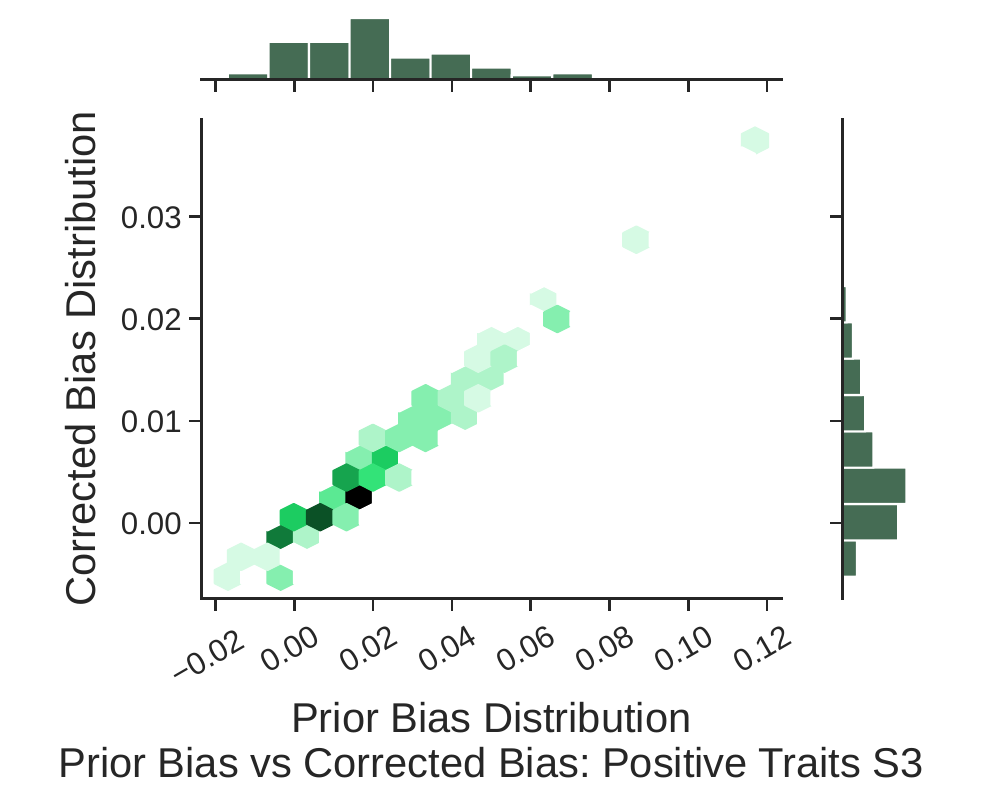}
        \end{subfigure}
        \begin{subfigure}{0.33\textwidth}
            \includegraphics[width=\textwidth, trim={0 0 0 0}, clip]{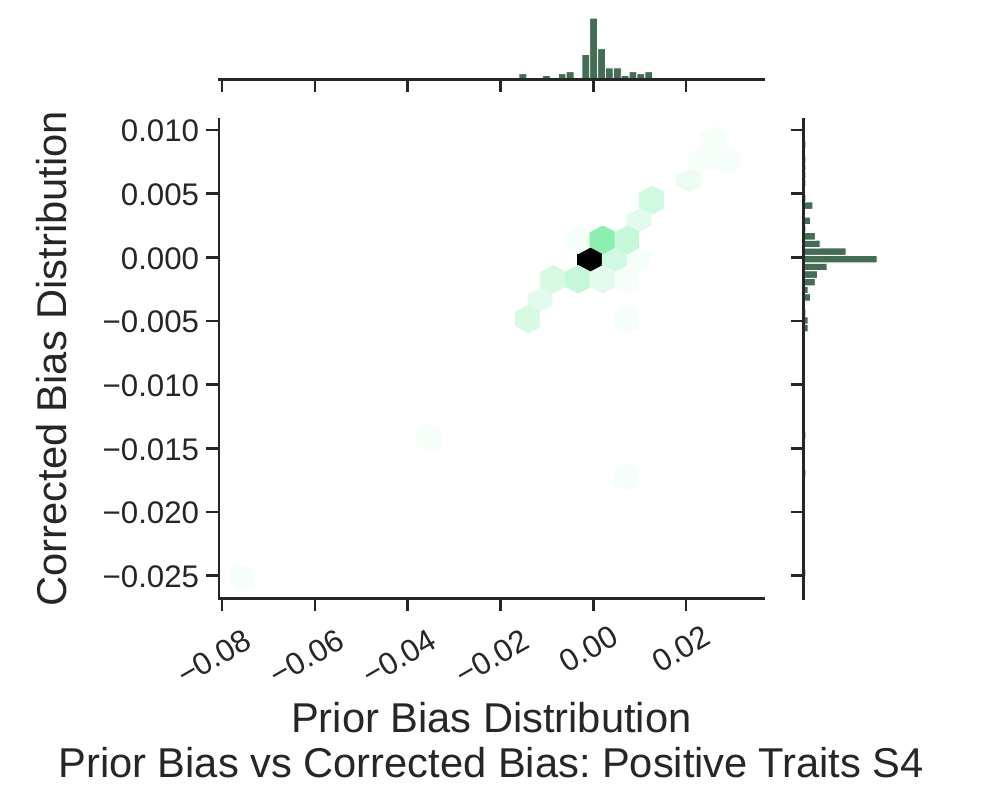}
        \end{subfigure}
        \begin{subfigure}{0.33\textwidth}
            \includegraphics[width=\textwidth, trim={0 0 0 0}, clip]{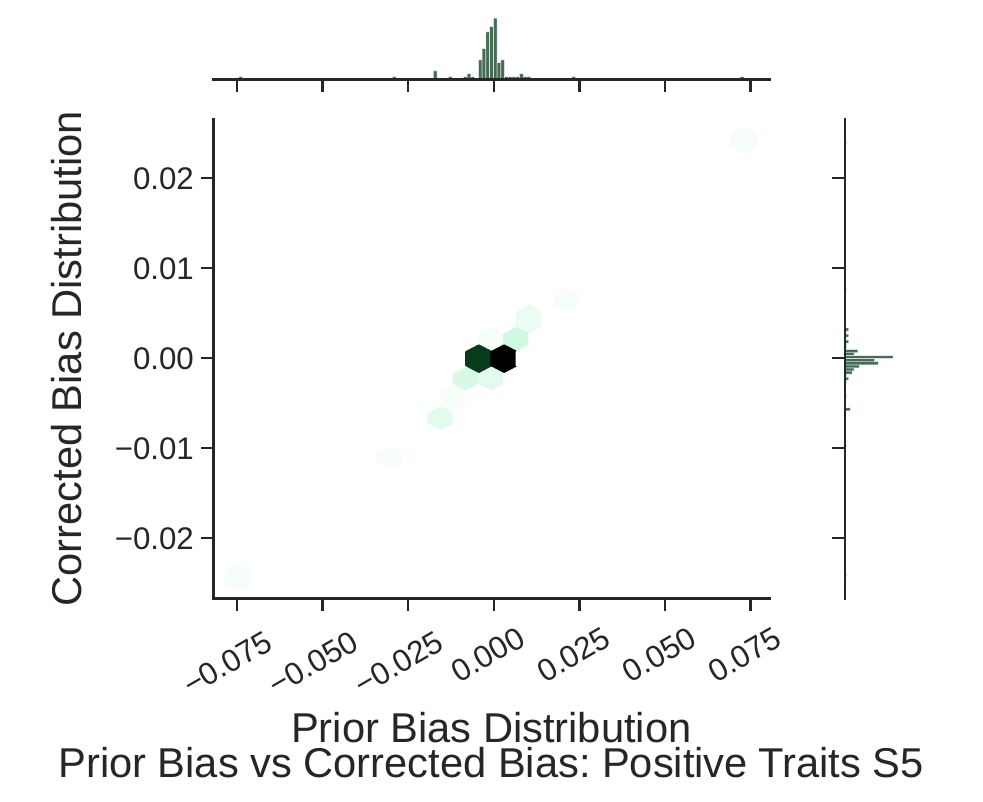}
        \end{subfigure}
        \caption{Prior Bias Score vs Corrected Bias Score plots for positive traits in \textbf{BanglaBERT} - Large Generator.}
    \end{subfigure}
    \\~\\
    \begin{subfigure}{\textwidth}
        \centering
        \begin{subfigure}{0.3\linewidth}
            \includegraphics[width=\linewidth, trim={0.5cm 0 1cm 0}, clip]{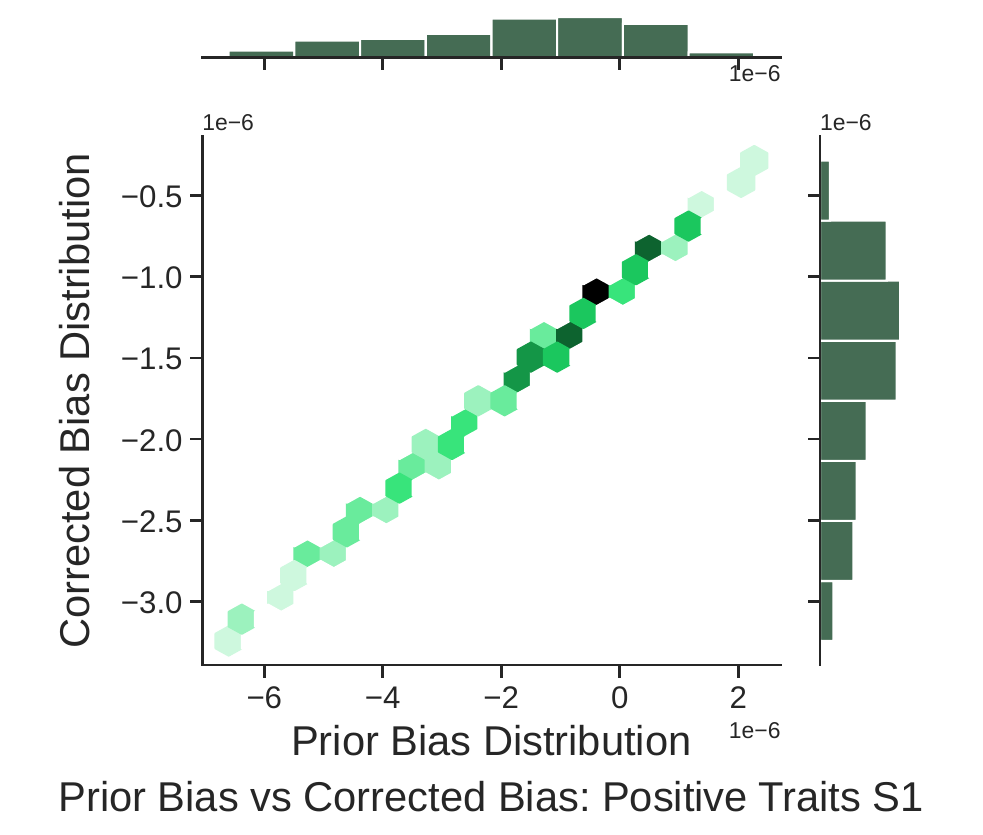}
        \end{subfigure}
        \begin{subfigure}{0.3\linewidth}
            \includegraphics[width=\linewidth, trim={0.5cm 0 1cm 0}, clip]{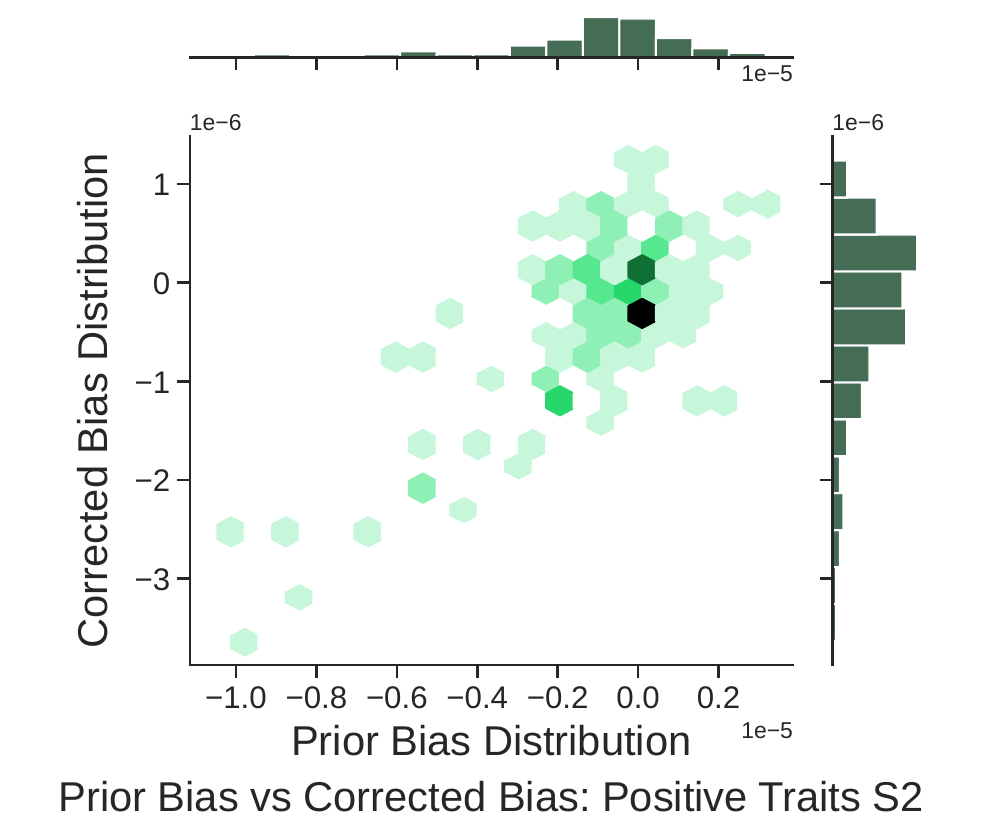}
        \end{subfigure}
        \begin{subfigure}{0.3\linewidth}
            \includegraphics[width=\linewidth, trim={0.5cm 0 1cm 0}, clip]{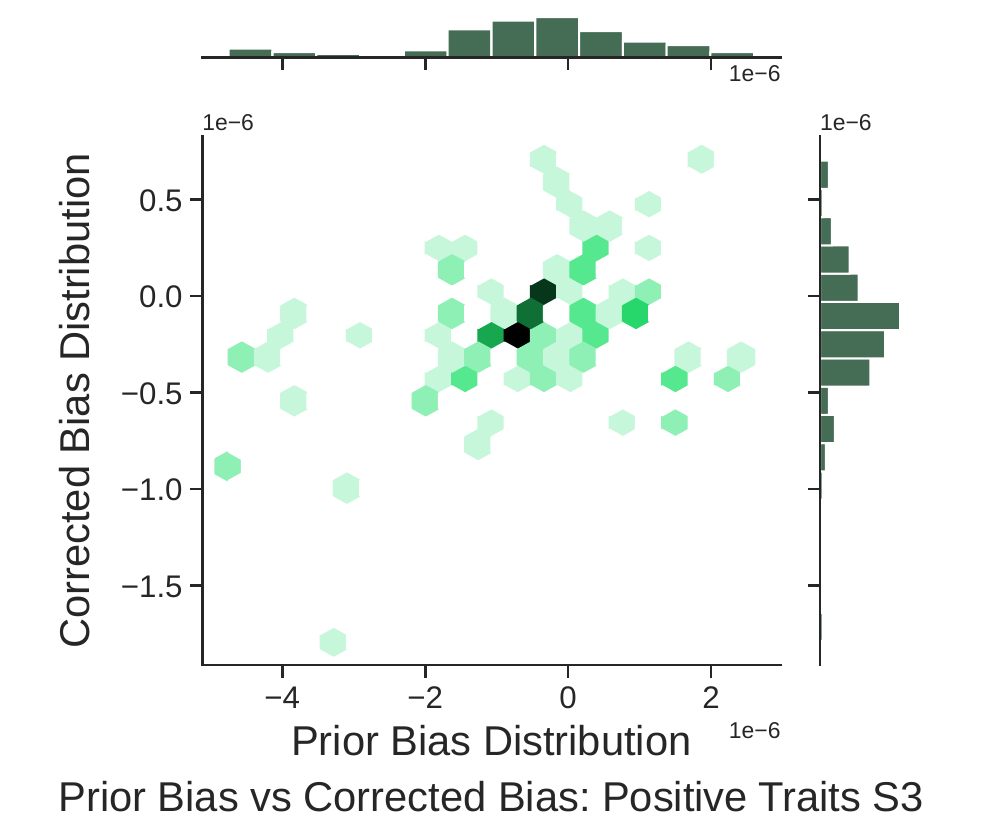}
        \end{subfigure}
        \begin{subfigure}{0.33\textwidth}
            \includegraphics[width=\textwidth, trim={0 0 0 0}, clip]{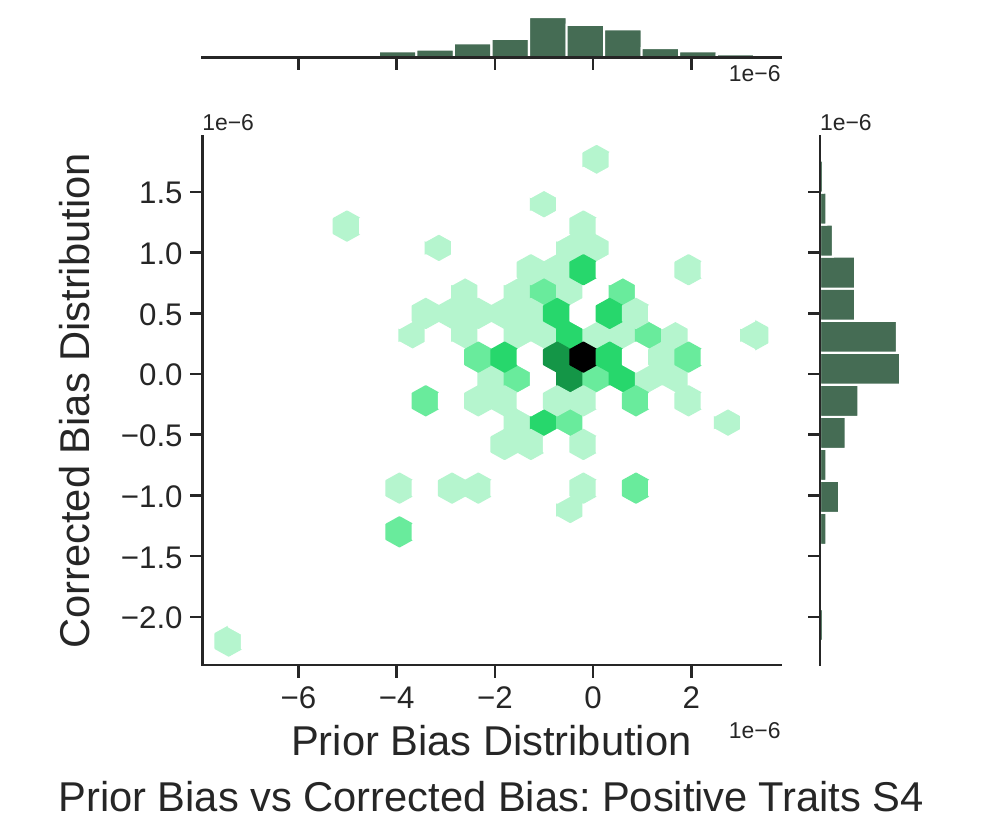}
        \end{subfigure}
        \begin{subfigure}{0.33\textwidth}
            \includegraphics[width=\textwidth, trim={0 0 0 0}, clip]{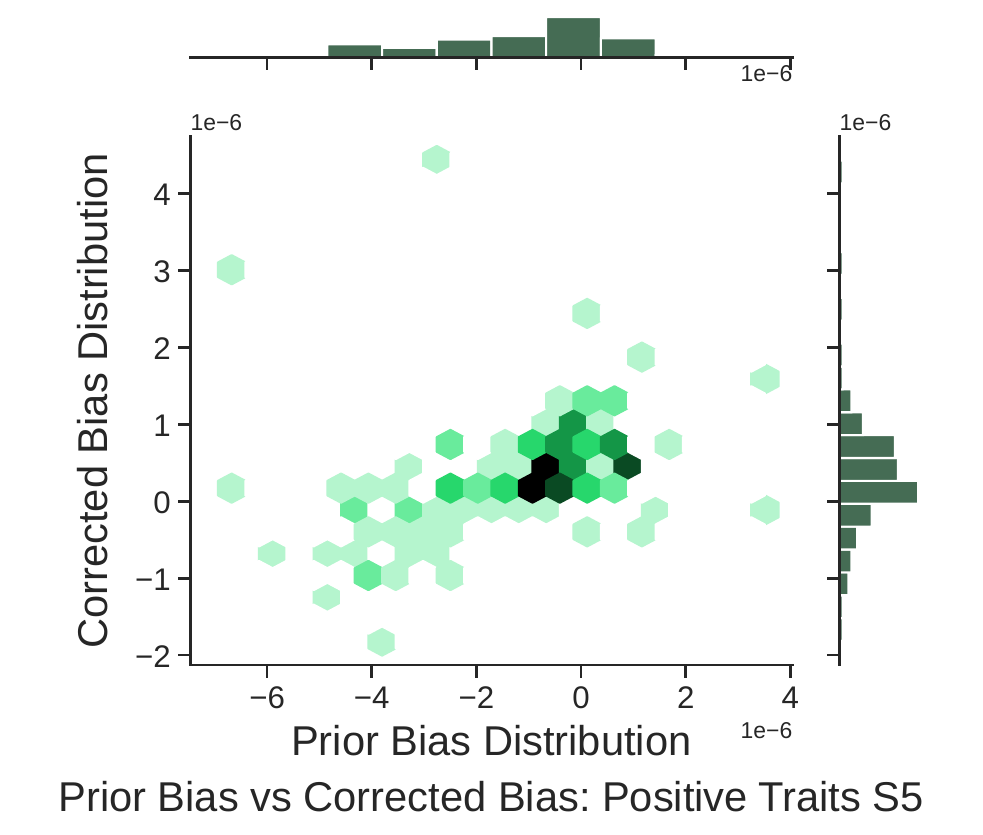}
        \end{subfigure}
        \caption{Prior Bias Score vs Corrected Bias Score plots for positive traits in \textbf{MuRIL} - Large (cased).}
    \end{subfigure}
\end{figure*}
\begin{figure*}[ht]\ContinuedFloat
    \begin{subfigure}{\textwidth}
        \centering
        \begin{subfigure}{0.3\linewidth}
            \includegraphics[width=\linewidth, trim={0.5cm 0 1cm 0}, clip]{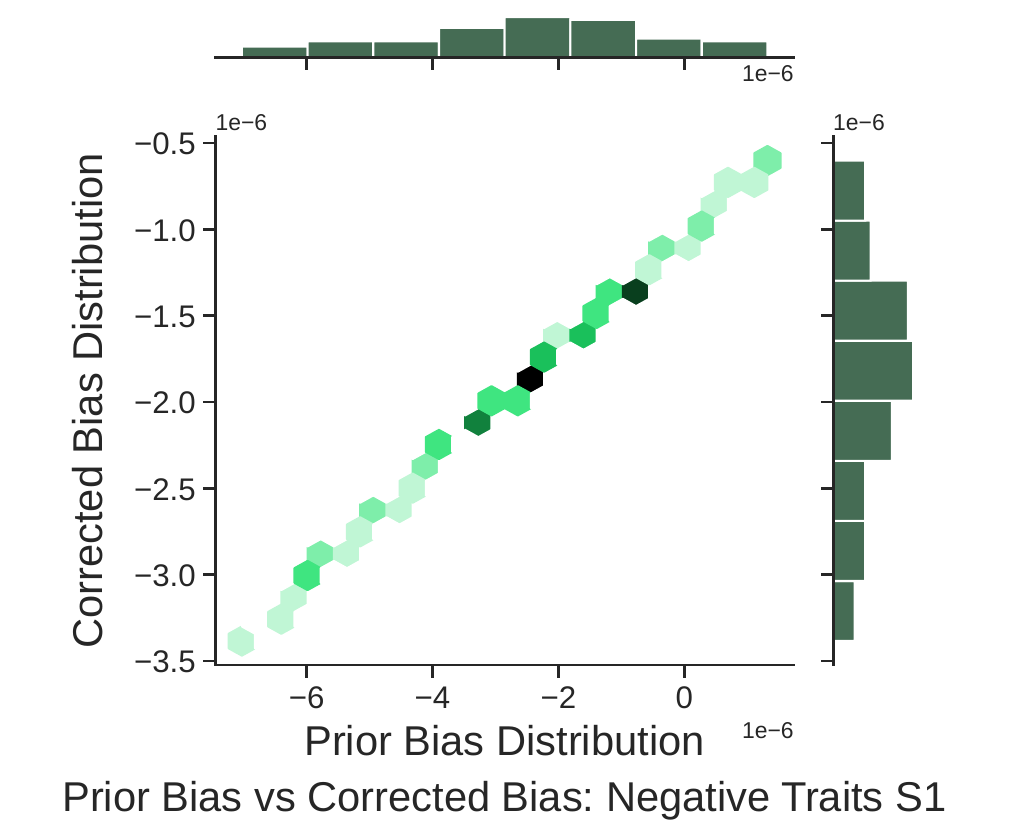}
        \end{subfigure}
        \begin{subfigure}{0.3\linewidth}
            \includegraphics[width=\linewidth, trim={0.5cm 0 1cm 0}, clip]{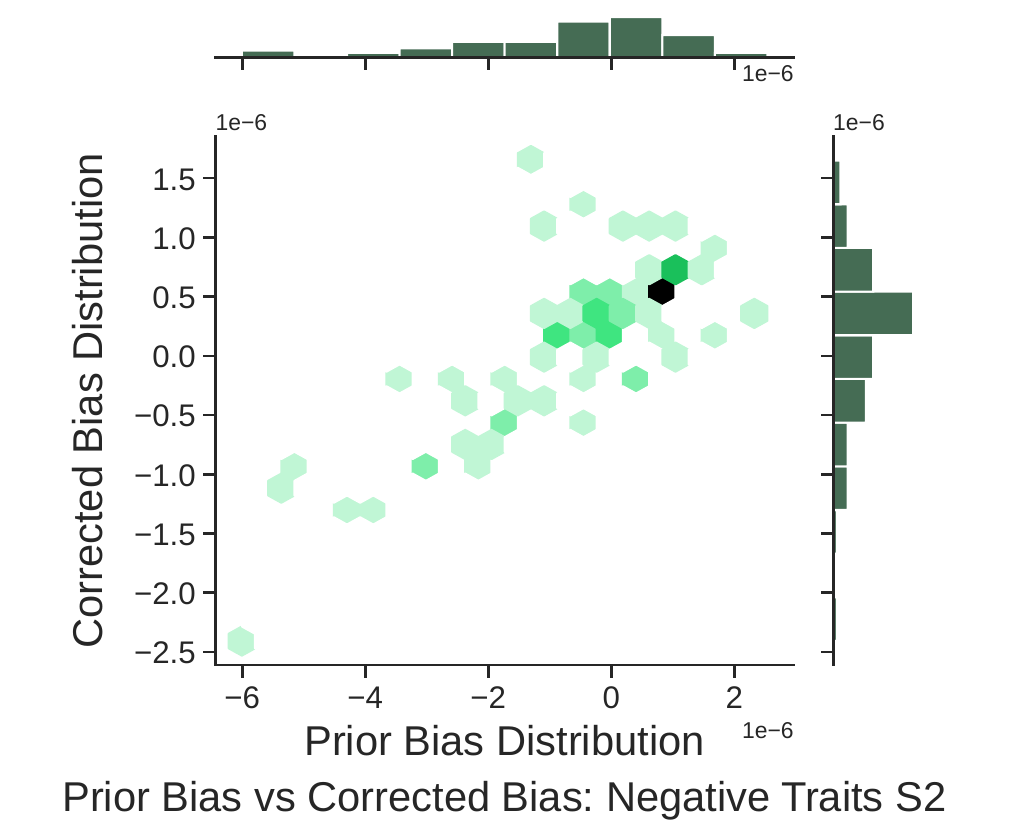}
        \end{subfigure}
        \begin{subfigure}{0.3\linewidth}
            \includegraphics[width=\linewidth, trim={0.5cm 0 1cm 0}, clip]{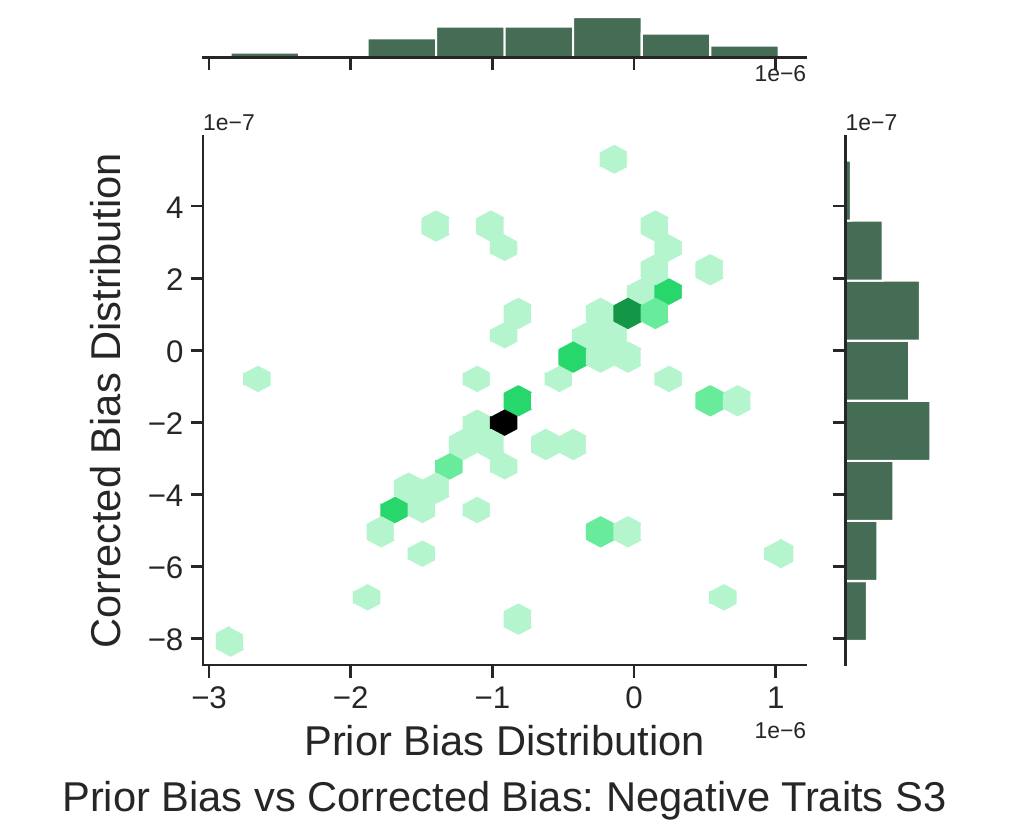}
        \end{subfigure}
        \begin{subfigure}{0.33\textwidth}
            \includegraphics[width=\textwidth, trim={0 0 0 0}, clip]{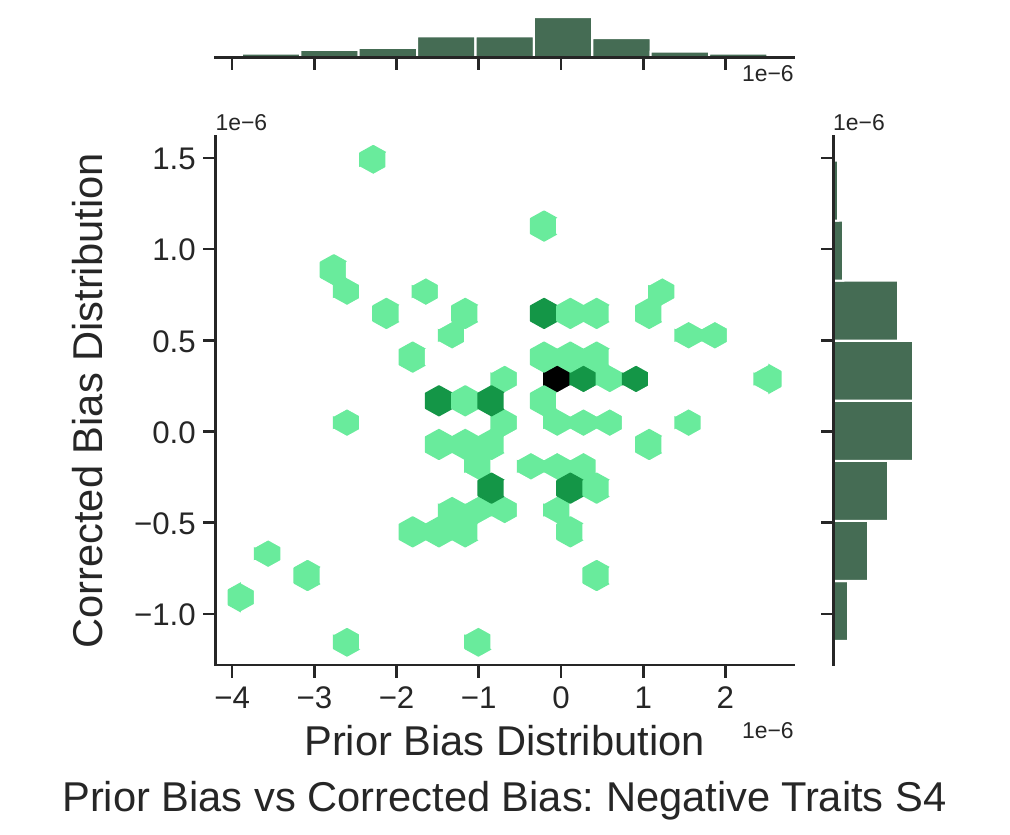}
        \end{subfigure}
        \begin{subfigure}{0.33\textwidth}
            \includegraphics[width=\textwidth, trim={0 0 0 0}, clip]{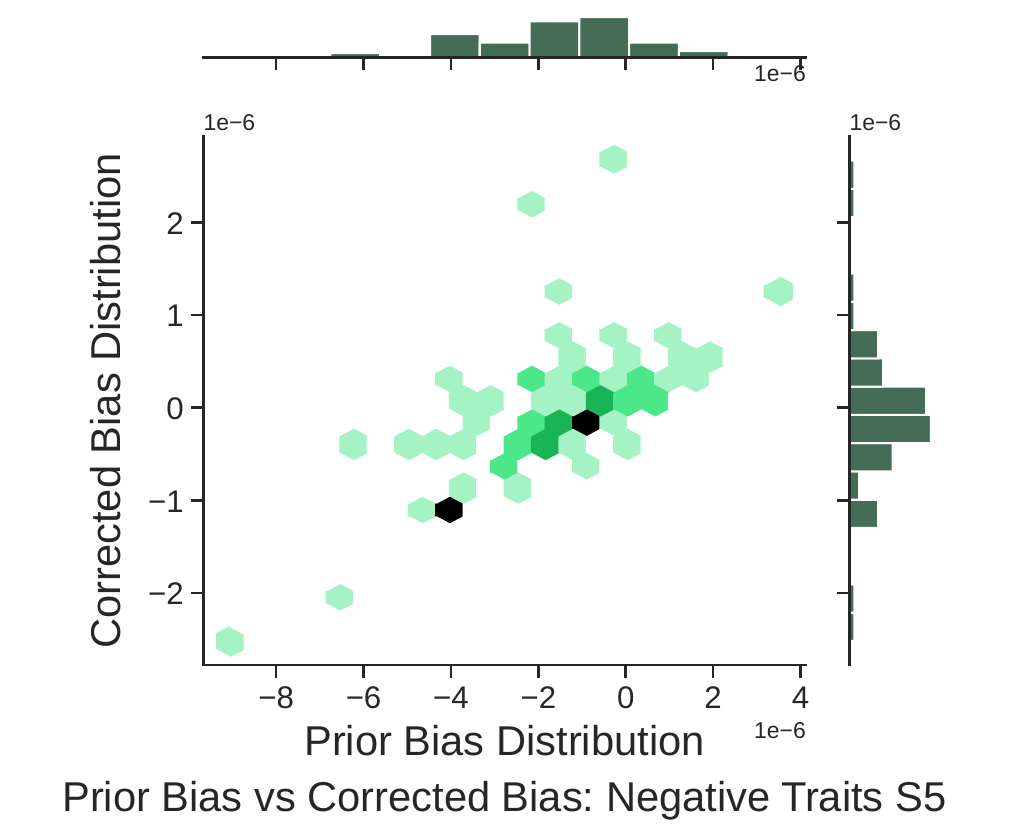}
        \end{subfigure}
        \caption{ Prior Bias Score vs Corrected Bias Score plots for negative traits in \textbf{MuRIL} - Large (cased).}
    \end{subfigure}
    \caption{A comparison between model behaviors for different sentence structures in Log Probability Bias Score Test.}
    \label{fig:bbgen_muril_hexbin}
\end{figure*}
\appendix